%% file: main.tex
\documentclass[10pt,twocolumn,letterpaper]{article}

\usepackage[pagenumbers]{iccv} % To force page numbers, e.g. for an arXiv version

\input{preamble}

\definecolor{cvprblue}{rgb}{0.21,0.49,0.74}
\usepackage[pagebackref,breaklinks,colorlinks,allcolors=cvprblue]{hyperref}
\usepackage{xcolor}
\usepackage{colortbl}
\usepackage{multirow}
\usepackage{caption}
\usepackage{tikz}
\def\paperID{8943} % *** Enter the Paper ID here
\def\confName{ICCV}
\def\confYear{2025}

\title{NexusSplats: Efficient 3D Gaussian Splatting in the Wild}

\author{Yuzhou Tang\quad
Dejun Xu\quad
Yongjie Hou\quad
Zhenzhong Wang\quad
Min Jiang${\footnotemark[2]}$\vspace{4pt}\\
School of Informatics, Xiamen University\vspace{3pt}\\
\url{https://nexus-splats.github.io/}\\
}

\begin{document}

\twocolumn[{%
	\renewcommand\twocolumn[1][]{#1}%
	\maketitle
	\vspace{-2em}
        \scalebox{0.98}{
        \begin{tikzpicture}
            \node[] (image) at (0,0) {};
            \def\imagePath{images/teasers/gt_1.jpg}
            \node[anchor=north west,inner sep=0,alias=image,line width=0pt] (image1) at (0,0) {\includegraphics[width=0.155\textwidth]{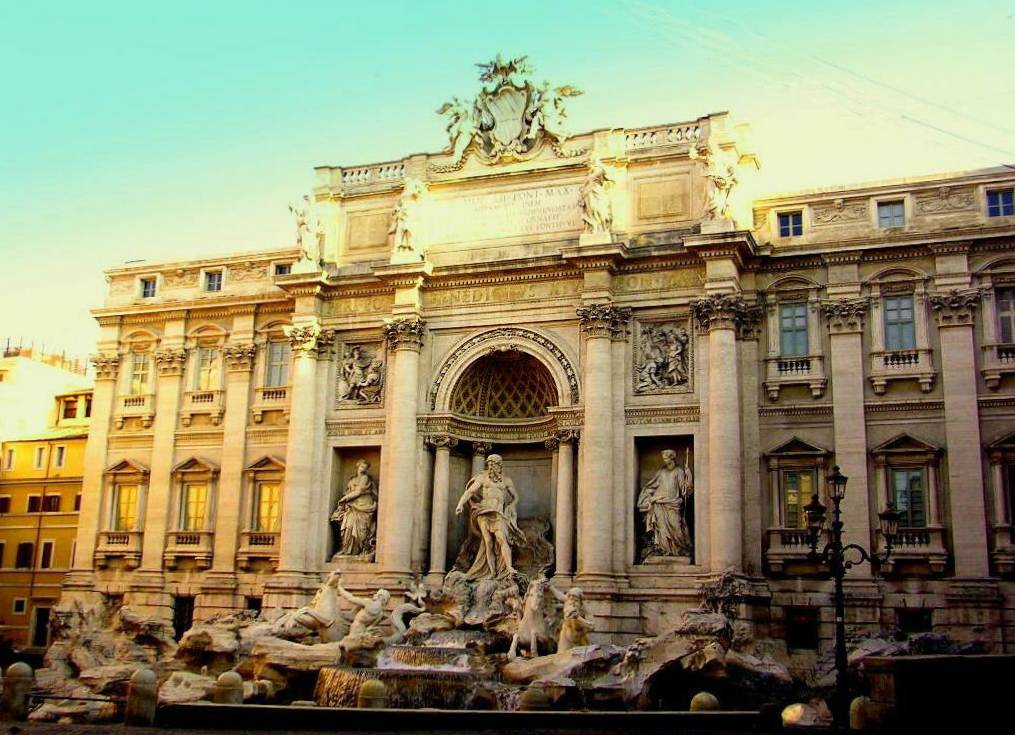}};
            \node[anchor=south west] (row-marker) at (image1.south west) {};
    
            \def\imagePath{images/teasers/teaser.png}
            \node[anchor=north west,inner sep=0,alias=image,line width=0pt,xshift=0.163\textwidth,yshift=0.001\textwidth] (image2) at (image1.north west) {\includegraphics[width=0.455\textwidth]{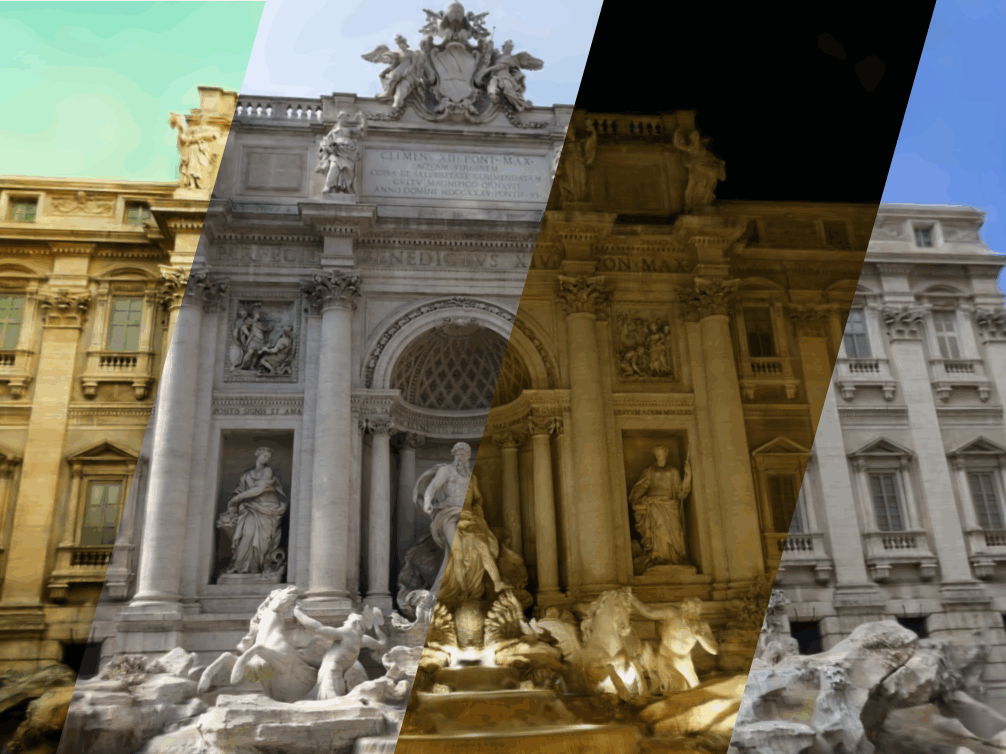}};
    
            \def\imagePath{images/teasers/metric.png}
            \node[anchor=north west,inner sep=0,alias=image,line width=0pt,xshift=0.46\textwidth,yshift=0.0015\textwidth] (image3) at (image2.north west) {\includegraphics[width=0.385\textwidth]{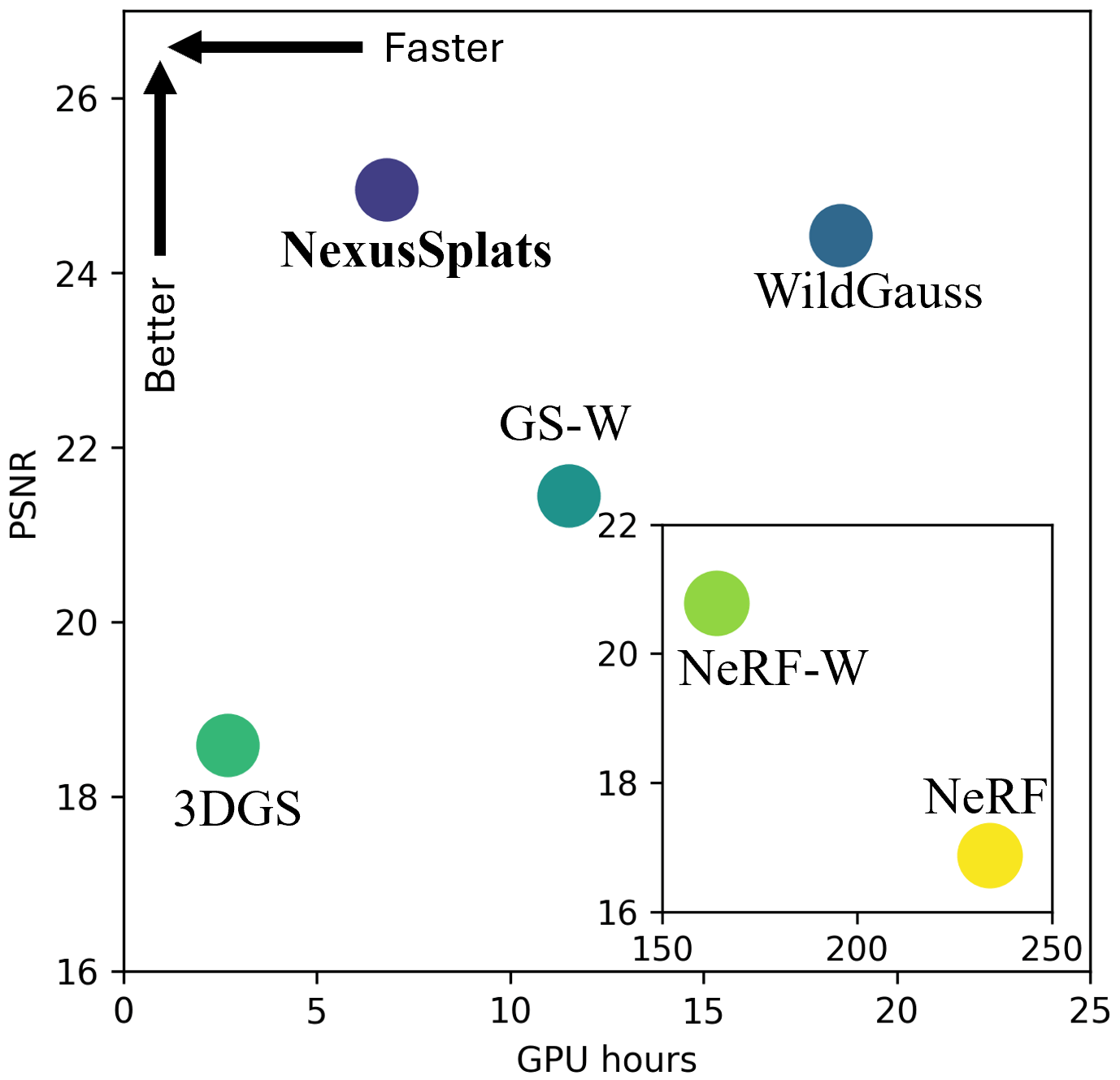}};
            
            \def\imagePath{images/teasers/gt_2.jpg}
            \node[anchor=north west,inner sep=0,alias=image,line width=0pt,yshift=-0.006\textwidth] (image4) at (row-marker.south west) {\includegraphics[width=0.155\textwidth]{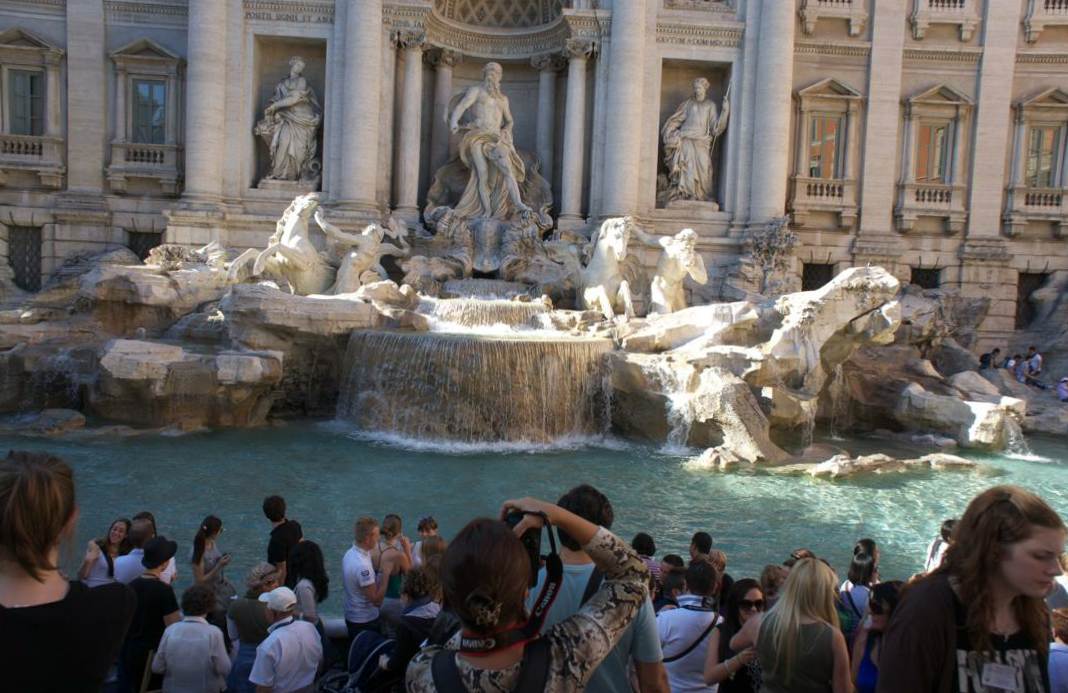}};
            \node[anchor=south west] (row-marker) at (image4.south west) {};
    
            \def\imagePath{images/teasers/gt_3.jpg}
            \node[anchor=north west,inner sep=0,alias=image,line width=0pt,yshift=-0.006\textwidth] (image5) at (row-marker.south west) {\includegraphics[width=0.155\textwidth]{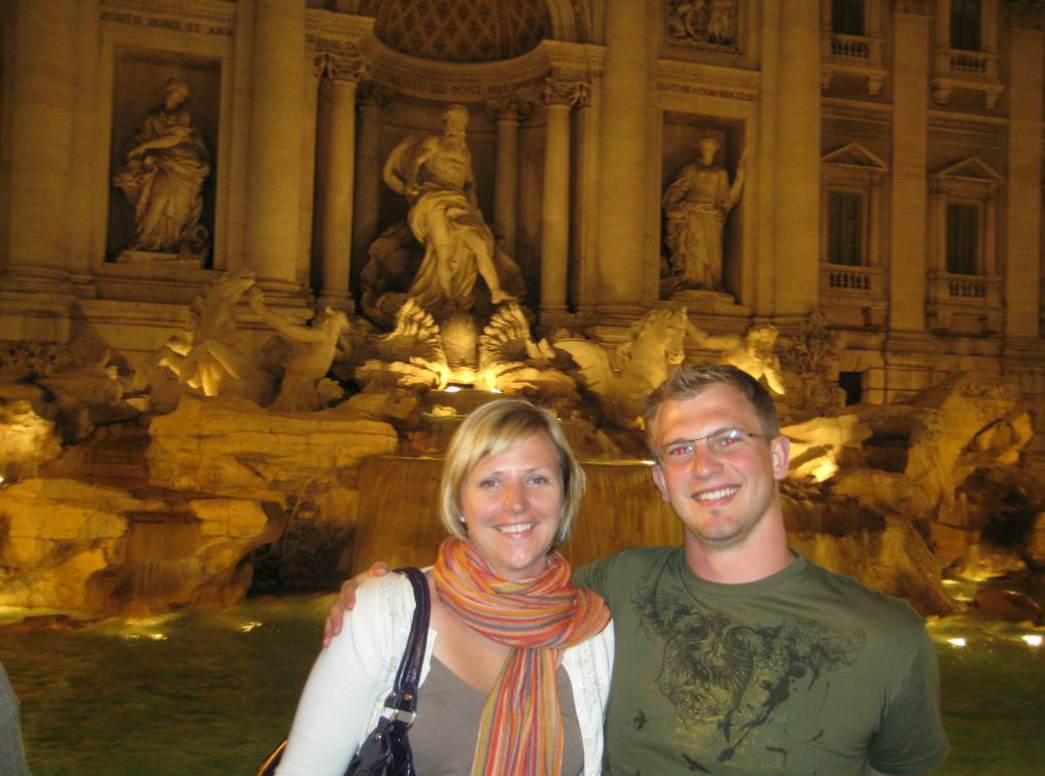}};
            \node[anchor=north, inner sep=2pt, yshift=15pt] at (image1.north) {(a) Photos};
            \node[anchor=north, inner sep=2pt, xshift=0.31\textwidth, yshift=15pt] at (image1.north) {(b) Renderings};
            \node[anchor=north, inner sep=2pt, xshift=0.755\textwidth, yshift=15pt] at (image1.north) {(c) Performance Comparison};
            
        \end{tikzpicture}
        }
        \vspace{-1.7em}
        \captionof{figure}{(a) Given photos from \textit{in-the-wild} scenarios, (b) our method decouples lighting conditions and eliminates occlusions, enabling steerable color mapping to diverse lighting conditions. (c) \textbf{NexusSplats} achieves state-of-the-art rendering quality, with a substantial training speed improvement over extensions of 3DGS.}
	\label{fig:teaser}
	\vspace{1em}
}]
% \maketitle

\footnotetext[2]{Corresponding author.}

\input{sec/0_abstract}
\input{sec/1_introduction}
\input{sec/2_related_work}
\input{sec/3_methods}
\input{sec/4_experiments}
\input{sec/5_conclusion}
{
    \small
    \bibliographystyle{ieeenat_fullname}
    \bibliography{main}
}

% WARNING: do not forget to delete the supplementary pages from your submission 
\input{sec/X_suppl}
% \input{sec/6_suppl}
% {
%     \small
%     \bibliographystyle{ieeenat_fullname}
%     \bibliography{main}
% }

\end{document}

%% file: preamble.tex
\definecolor{tabfirst}{rgb}{1, 0.7, 0.7} % red
\definecolor{tabsecond}{rgb}{1, 0.85, 0.7} % orange
\definecolor{tabthird}{rgb}{1, 1, 0.7} % yellow

\makeatletter
\newcommand{\Rmnum}[1]{\expandafter\@slowromancap\romannumeral #1@}
\makeatother

%% file: sec/0_abstract.tex
\begin{abstract}
% \vspace{-0.3em}
Photorealistic 3D reconstruction of unstructured real-world scenes remains challenging due to complex illumination variations and transient occlusions. Existing methods based on Neural Radiance Fields (NeRF) and 3D Gaussian Splatting (3DGS) struggle with inefficient light decoupling and structure-agnostic occlusion handling.
To address these limitations, we propose NexusSplats, an approach tailored for efficient and high-fidelity 3D scene reconstruction under complex lighting and occlusion conditions. 
In particular, NexusSplats leverages a hierarchical light decoupling strategy that performs centralized appearance learning, efficiently and effectively decoupling varying lighting conditions. Furthermore, a structure-aware occlusion handling mechanism is developed, establishing a nexus between 3D and 2D structures for fine-grained occlusion handling.
Experimental results demonstrate that NexusSplats achieves state-of-the-art rendering quality and reduces the number of total parameters by 65.4\%, leading to 2.7$\times$ faster reconstruction. 

\end{abstract}

%% file: sec/1_introduction.tex
\section{Introduction}
\label{sec:intro}

Photorealistic 3D scene reconstruction from unstructured image collections remains a cornerstone for applications ranging from virtual reality to autonomous navigation \cite{martin2021nerfw,chen2022hallucinated}. While Neural Radiance Fields (NeRFs) \cite{mildenhall2021nerf,barron2022mip,tancik2023nerfstudio,rematas2022urban} achieve impressive view synthesis in controlled settings, their volumetric rendering struggles with \textit{in-the-wild} scenarios characterized by illumination variations (e.g. weather changes) and transient occlusions (e.g. moving pedestrians). NeRF-W \cite{martin2021nerfw} pioneers light decoupling and occlusion handling through image-specific appearance and transient embeddings combined with color mapping and 2D uncertainty estimation. Follow-up works \cite{chen2022hallucinated,rudnev2022relighting,li2023nerfms,yang2023cross,kassab2023refinedfields} enhance the appearance and uncertainty modeling by introducing various image feature extractors, such as CNN and U-Net. However, these methods inherit NeRF's prohibitive computational cost, requiring hours of training per scene.

The emergence of 3D Gaussian Splatting (3DGS) \cite{kerbl20233d,yu2024mipsplatting,yu2024gaussianopcacityfields} revolutionizes the field with explicit 3D Gaussians and differentiable rasterization \cite{zwicker2002ewa}, enabling faster reconstruction with real-time rendering. Recent adaptations \cite{kulhanek2024wildgaussians,zhang2024gaussianw,xu2024splatfacto,xu2024wildgs} extend 3DGS to \textit{in-the-wild} scenes by introducing per-Gaussian appearance embeddings and scattered color mapping to decouple lighting conditions from each explicit Gaussian primitive. To handle transient occlusions, these methods remain their focus on advanced image feature extractors for uncertainty modeling.

Despite progress in improving rendering quality, two fundamental limitations persist:
(1) \textbf{Inefficient light decoupling.} Representing outdoor scenes with millions of 3D Gaussians, current methods attach independent appearance embeddings to the substantial number of Gaussians. This scattered appearance learning increases the total number of parameters and slows training, while simplified alternatives like positional encodings \cite{dahmani2024swag} sacrifice texture fidelity due to insufficient learning of detailed appearance features.
(2) \textbf{Structure-agnostic occlusion handling.} Existing 2D-centric occlusion detection fails to account for the inherent structural discrepancies between 2D input images and the 3D representation, limiting the accuracy in capturing occlusions. Besides, clear scene boundaries where no occlusions appear are frequently misclassified as occlusions by existing methods. Excluding clear areas while retaining uncaptured occlusions leads to incomplete scene reconstruction with artifacts and shadows.

The key to addressing these issues lies in two key perspectives:
(1) Construct an efficient structure that learns sufficient appearance details while scaling down the parameters. 
(2) Build a nexus between 3D and 2D structures to enable more accurate occlusion captures.
Based on this, we propose \textbf{NexusSplats} that performs hierarchical light decoupling and structure-aware occlusion handling.

To enable hierarchical light decoupling, we introduce the hierarchical and region-aware 3D representation \cite{lu2024scaffold} to organize the 3D Gaussians into \textit{nexus kernels} --- dynamic 3D primitives tailored for \textit{in-the-wild} scene reconstruction. In particular, each \textit{nexus kernel} learns a shared appearance embedding and coordinates color mapping for its managed Gaussians, allowing for centralized appearance learning while scaling down the number of appearance embeddings. This design maintains sufficient detail learning and ensures rendering quality.
Regarding occlusion handling, a shared uncertainty embedding is also assigned for each kernel to predict the 3D uncertainties of the associated Gaussians. These 3D uncertainties are then projected onto image planes via tile rasterization, propagating to 2D uncertainty masks. To further correct misidentifications on clear scene boundaries, we apply a boundary-aware refinement to the 2D uncertainty masks. Combined with the 2D semantic feature loss \cite{ren2024nerfonthego,kulhanek2024wildgaussians}, our uncertainty propagation mechanism bridges 3D uncertainties with 2D semantic features, improving the accuracy of capturing true occlusions.
Extensive experiments demonstrate that our method not only significantly reduces redundant parameters but also presents finer textures and a more accurate removal of occlusions.

In summary, our key contributions are as follows:
\begin{itemize}
    \item We develop a hierarchical light decoupling strategy that performs centralized appearance learning, efficiently and effectively decoupling varying lighting conditions.
    \item We design a structure-aware occlusion handling mechanism that establishes a nexus between 3D and 2D structures, improving the accuracy of occlusion captures.
    \item Experimental results demonstrate that NexusSplats achieves state-of-the-art rendering quality and reduces total parameters by 65.4\%, which leads to 2.7$\times$ faster reconstruction.
\end{itemize}

%% file: sec/2_related_work.tex
\section{Related Work}
\subsection{Novel View Synthesis}
NeRFs \cite{mildenhall2021nerf,barron2022mip,tancik2023nerfstudio,rematas2022urban} revolutionize 3D scene reconstruction through continuous volumetric representations, enabling photorealistic novel view synthesis. However, their reliance on ray tracing and MLP-based rendering imposes prohibitive computational costs \cite{chen2019learning,mescheder2019occupancy,park2019deepsdf}.
Subsequent works accelerate NeRFs through enhanced scene representations \cite{chen2022tensorf,fridovich2022plenoxels,kerbl20233d,liu2020neural,muller2022instant,sun2022direct}, where 3DGS \cite{kerbl20233d,yu2024mipsplatting,yu2024gaussianopcacityfields} excels in its remarkable efficiency without compromising rendering quality by combining explicit 3D Gaussians with differentiable rasterization. While follow-up works \cite{lu2024scaffold,lin2024vastgaussian,fan2023lightgaussian,lee2024compact,morgenstern2023compact,navaneet2023compact3d,niedermayr2024compressed} further compress 3DGS representations for settling large-scale scenes, they remain inadequate in handling real-world challenges like complex illumination variations and transient occlusions. In this paper, we investigate scene representations tailored for such challenging conditions.

%-------------------------------------------------------------------------
\subsection{Lighting Condition Decoupling}
To handle illumination variations, NeRF-based methods \cite{martin2021nerfw,chen2022hallucinated,rudnev2022relighting,li2023nerfms,yang2023cross,kassab2023refinedfields} typically employ image-specific embeddings and neural feature extractors to disentangle appearance changes. However, these approaches inherit NeRF's computational inefficiency, resulting in prohibitive training and rendering times. Recent efforts \cite{zhang2024gaussianw,kulhanek2024wildgaussians,xu2024splatfacto,xu2024wildgs} integrate appearance conditioning with 3DGS by introducing per-Gaussian appearance embeddings and scattered color mapping. Though proven to be effective, optimizing such a substantial number of extra parameters significantly increases computational costs. While SWAG \cite{dahmani2024swag} proposes positional encodings to replace per-Gaussian embeddings for higher efficiency, this strategy sacrifices texture fidelity due to limited appearance modeling capacity. Our approach strikes a novel balance between efficiency and quality by introducing a hierarchical light decoupling strategy that maintains high-fidelity rendering without excessive parameters.

%-------------------------------------------------------------------------
\subsection{Transient Occlusion Handling}
To mitigate the impact of transient occlusions, such as moving pedestrians and vehicles, pioneering works \cite{martin2021nerfw,chen2022hallucinated,yang2023cross,sabour2023robustnerf} either predict 2D uncertainty masks or leverage loss controls based on image feature extractors (e.g. CNN and U-Net) to capture and eliminate occlusions, thereby preventing their models from learning irrelevant objects from training images. Subsequent methods \cite{ren2024nerfonthego,kulhanek2024wildgaussians} enhance occlusion detection using advanced semantic feature extractors like DINO \cite{caron2021emerging,oquab2023dinov2}. 
However, their reliance on 2D-informed guidance can lead to inconsistencies with the underlying 3D scene geometry, making them more prone to occasional false or missed occlusion identifications. In addition, specialized 3DGS variants for occlusion handling \cite{sabourgoli2024spotlesssplats,lin2024hybridgs} primarily focus on small-scale indoor scenarios, which involve occlusion challenges but lack drastic illumination variations in real-world outdoor environments.
In contrast, our method introduces a structure-aware occlusion handling framework designed for large outdoor scenes, where complex illumination changes and transient objects coexist. This dual consideration of lighting and occlusion distinguishes our approach from existing solutions that address transient occlusions in isolation.

%% file: sec/3_methods.tex
\section{Methods}
\begin{figure*}[ht]
    \centering
    \includegraphics[width=1.\linewidth]{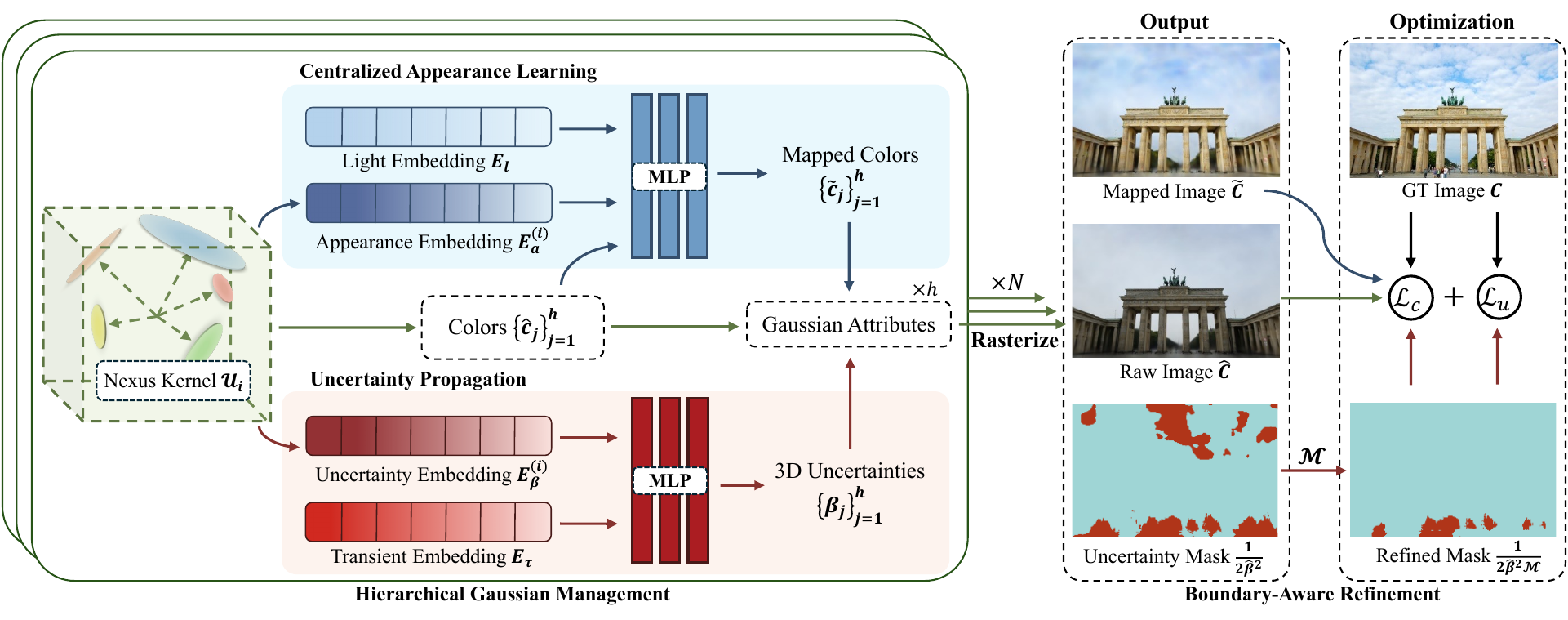}
    \vspace{-2em}
    \caption{\textbf{Overview of NexusSplats.} Our framework operates in three stages: \textit{First}, the hierarchical Gaussian management (\Cref{sec:hierarchical management}) organizes 3D Gaussians into dynamic \textit{nexus kernels}, which generate Gaussian attributes and perform centralized appearance learning (\Cref{sec:centralized appearance learning}) and uncertainty propagation (\Cref{sec:uncertainty propagation}). \textit{Second}, a raw image $\hat{\mathbf{C}}$, a mapped image $\tilde{\mathbf{C}}$, and an uncertainty mask $\frac{1}{2\hat{\mathbf{\beta}}^2}$ are rendered through tile rasterization. \textit{Third}, the boundary-aware refinement (\Cref{sec:boundary-aware refinement}) corrects misclassified scene boundaries. The system optimizes via a combination of color loss $\mathcal{L}_c$ and uncertainty loss $\mathcal{L}_u$.}
    \label{fig:overview}
    \vspace{-0.5em}
\end{figure*}

\Cref{fig:overview} illustrates our approach, NexusSplats, an approach addressing inefficient light decoupling and structure-agnostic occlusion handling in unstructured 3D scene reconstruction. Our approach comprises two key components:
(1) \textbf{Hierarchical Light Decoupling} (\Cref{sec:light decoupling}) that leverages \textit{nexus kernels} to hierarchically manage 3D Gaussians and perform centralized appearance learning for efficient and effective light condition decoupling.
(2) \textbf{Structure-Aware Occlusion Handling} (\Cref{sec:occlusion handling}) that bridges 3D uncertainties with 2D semantic features via uncertainty propagation and incorporates the boundary-aware refinement to mitigate misclassifications at scene boundaries.
Using a combination of color loss and uncertainty loss, NexusSplats enables delicate scene reconstructions with steerable lighting conditions via unified optimization (\Cref{sec:optimization}).
In this section, we detail each component of NexusSplats, elaborating on how they collectively contribute to efficient and high-quality 3D scene reconstruction under complex conditions.

%-------------------------------------------------------------------------
%-------------------------------------------------------------------------
\subsection{Preliminaries}
\label{sec:preliminaries}
In this section, we briefly review 3DGS \cite{kerbl20233d} and its extension Scaffold-GS \cite{lu2024scaffold}, which jointly serve as a basis of our approach. These methods combine explicit 3D Gaussian representations with differentiable rasterization for efficient scene reconstruction.

\paragraph{3D Gaussian Splatting} represents scenes using anisotropic 3D Gaussians initialized from Structure-from-Motion (SfM) point clouds \cite{schonberger2016structure}. Each Gaussian $\mathcal{G}(\mathbf{x})$ is defined by:
\begin{equation}
  \mathcal{G}(\mathbf{x})=e^{-\frac{1}{2}(\mathbf{x}-\mathbf{\mu})^\top\mathbf{\Sigma}^{-1}(\mathbf{x}-\mathbf{\mu})} ,\quad\mathbf{\Sigma}=\mathbf{R}\mathbf{S}\mathbf{S}^\top\mathbf{R}^\top,
  \label{eq:gaussian}
\end{equation}
where $\mathbf{\mu}$ is the centroid position of a 3D Gaussian, and $\mathbf{\Sigma}$ encodes anisotropic covariance via a scaling matrix $\mathbf{S}$ and a rotation matrix $\mathbf{R}$. Each Gaussian also stores the view-dependent color $\hat{\mathbf{c}}$, modeled by spherical harmonics, and an opacity $\alpha$.

Unlike traditional volumetric methods, 3DGS uses tile-based rasterization \cite{zwicker2002ewa} for real-time rendering. Each 3D Gaussian $\mathcal{G}(\mathbf{x})$ is projected onto image planes, resulting in a 2D Gaussian $\mathcal{G}'(\mathbf{x}')$. The 2D Gaussians are then sorted and splatted via $\alpha$-blending:
\begin{equation}
  \hat{\mathbf{C}}(\mathbf{x}')=\sum_{i\in N}\hat{\mathbf{c}}_i\sigma_i\prod_{j=1}^{i-1}(1-\sigma_j),\quad\sigma_i=\alpha_i\mathcal{G}_i'(\mathbf{x}') ,
  \label{eq:splatting}
\end{equation}
where $\sigma_i$ denotes the contribution weight of the $i$-th Gaussian to pixel $\mathbf{x}'$. With the differentiable property, all attributes of 3D Gaussians are learnable and optimized end-to-end during training.

\paragraph{Scaffold-GS} introduces a hierarchical representation to address parameter redundancy in large-scale scenes. 
Instead of directly optimizing a large, unstructured set of Gaussians, Scaffold-GS organizes the scene using a sparse grid of anchor points $\{\mathcal{V}_i\}$, initialized from SfM point clouds and dynamically generating $h$ neural Gaussians per anchor:
\begin{equation}
    \{\mathbf{\mu}_j\}_{j=1}^h = \mathbf{x}_i + \{\mathcal{O}_j\}_{j=1}^h \cdot l_v,
    \label{eq:neural_gaussian_position}
\end{equation}
where $\mathbf{x}_i$ is the position of anchor $\mathcal{V}_i$, and $\{\mathcal{O}_j\}$ and $l_v$ are learnable offsets and scaling factors.

To generate the attributes of each neural Gaussian, Scaffold-GS employs respective MLPs for prediction. Notably, the spherical harmonics with dynamic color channels are replaced with neural networks with fixed channels, consistent with other Gaussian attributes. Specifically, given viewing position $\mathbf{x}_v$, anchor $\mathcal{V}_i$ predicts the colors $\{\hat{\mathbf{c}}_j\}_{j=1}^{h}$ of neural Gaussians by:
\begin{equation}
    \{\hat{\mathbf{c}}_j\}_{j=1}^{h}=F_c(\mathbf{f}_{v},\vec{d}_{iv}),\quad\vec{d}_{iv}=\frac{\mathbf{x}_i-\mathbf{x}_v}{\|\mathbf{x}_i-\mathbf{x}_v\|_2},
    \label{eq:neural_gaussian_color}
\end{equation}
where $F_{c}$ is the color MLP shared by all anchors and the viewing direction $\vec{d}_{iv}$ ensures viewing consistency. Analogous MLPs, $F_\alpha$, $F_q$, and $F_s$, are used for opacity, covariance matrix, and scaling prediction. Finally, a unique optimization process is developed to prune redundant anchors with associated neural Gaussians and reduce total parameters.

%-------------------------------------------------------------------------
\subsection{Hierarchical Light Decoupling}
\label{sec:light decoupling}
This section introduces our hierarchical light decoupling strategy that consists of hierarchical Gaussian management and centralized appearance learning. This design effectively reduces parameter redundancy while achieving superior texture fidelity under varying illumination.

\subsubsection{Hierarchical Gaussian Management}
\label{sec:hierarchical management}
The core objective of hierarchical light decoupling is to establish an efficient representation that learns sufficient appearance features across 3D Gaussians while minimizing parameter overhead. A natural strategy involves reducing the number of appearance embeddings by grouping spatially adjacent Gaussians with similar appearance characteristics. However, roughly clustering Gaussian primitives based on their positions and colors suffers from two critical limitations.

First, leveraging a static clustering approach, where 3D Gaussians are roughly grouped based on their initial positions and colors, fails to capture the dynamic changes in Gaussian attributes (e.g. position, opacity, color) during training, leading to inconsistent appearance groupings. Second, introducing dynamic clustering requires frequent re-clustering which incurs prohibitive computational costs due to pairwise similarity calculations across millions of Gaussians.

To address these challenges, we draw inspiration from the hierarchical organization strategy of Scaffold-GS \cite{lu2024scaffold} and utilize it for Gaussian management.
Expanding on the concept of anchor points, we introduce \textit{nexus kernels}, which inherit the neural Gaussian generation capability and learn detailed appearance features in a centralized manner.
In this way, each \textit{nexus kernel} ensures the appearance similarities of the neural Gaussians it spawns, paving the way for centralized appearance learning.

\subsubsection{Centralized Appearance Learning}
\label{sec:centralized appearance learning}
Equipped with the shared appearance embedding $\mathbf{\varepsilon}_a^{(i)}$, each \textit{nexus kernel} $\mathcal{U}_i$ learns centralized appearance features and coordinates color mapping for its $h$ constituent Gaussians through a shared lighting-sensitive MLP $F_\theta$:
\begin{equation}
  \{\tilde{\mathbf{c}}_j\}_{j=1}^h= F_\theta(\{\hat{\mathbf{c}}_j\}_{j=1}^h, \mathbf{\varepsilon}_a^{(i)}, \mathbf{\varepsilon}_l^{(v)}, \vec{d}_{iv}) ,
  \label{eq:color-mapping}
\end{equation}
where $\mathbf{\varepsilon}_l^{(v)}$ is the view-specific light embedding from input image $C_v$, following previous literature \cite{martin2021nerfw,kulhanek2024wildgaussians,dahmani2024swag}. 
To maintain view consistency after color mapping, we also input the viewing direction $\vec{d}_{iv}$ into $F_\theta$, aligned with the raw color prediction process (\cref{eq:neural_gaussian_color}). Through tile rasterization, both raw colors $\hat{\mathbf{c}}$ and mapped colors $\tilde{\mathbf{c}}$ are projected onto image planes for loss computation.

By binding color transformations to kernel-level appearance embeddings rather than individual Gaussians, our method not only accelerates convergence through parameter sharing but also enhances representational capacity by facilitating appearance embeddings $\{\mathbf{\varepsilon}_a^{(i)}\}$ to encode material-aware illumination responses and structural coherence. This centralized paradigm contrasts sharply with conventional scattered embedding approaches, where isolated per-Gaussian parameters struggle to capture non-local lighting effects. Critically, the fixed-channel color expression provides a more stable optimization landscape compared to spherical harmonics with dynamic channels that are incompatible with neural network-based color mapping. 
% The enforced spatial-semantic grouping within \textit{nexus kernels} ensures consistent material property propagation across neighboring Gaussians, yielding smooth texture transitions under complex illumination shifts.

%-------------------------------------------------------------------------
\subsection{Structure-Aware Occlusion Handling}
\label{sec:occlusion handling}
This section elaborates on our structure-aware occlusion handling framework that bridges 2D uncertainty estimation with 3D representations to improve the accuracy of occlusion captures. In particular, our design leverages the 3D uncertainty estimation that grounds occlusion reasoning in geometric primitives and boundary-aware refinement that resolves semantic inconsistencies at scene peripheries.

\subsubsection{Uncertainty Propagation}
\label{sec:uncertainty propagation}
Central to this design is the prediction of 3D uncertainties $\{\mathbf{\beta}_i\}$ for individual Gaussians through \textit{nexus kernels}. Learning an uncertainty embedding $\mathbf{\varepsilon}_\beta^{(i)}$, each \textit{nexus kernel} $\mathcal{U}_i$ produces per-Gaussian confidence scores through a shared MLP $F_\beta$:
\begin{equation}
  \{\mathbf{\beta}_j\}_{j=1}^h= F_\beta(\mathbf{\varepsilon}_\beta^{(i)}, \mathbf{\varepsilon}_\tau^{(v)}) ,
  \label{eq:uncertainty predicting}
\end{equation}
where $\mathbf{\varepsilon}_\tau$ is the transient embedding for recording image-specific occlusion patterns.
These 3D uncertainties $\{\mathbf{\beta}_i\}$ are projected onto 2D image planes via tile rasterization:
\begin{equation}
  \hat{\mathbf{\beta}}(\mathbf{x}')=\sum_{j\in h\times N}\beta_j\sigma_j\prod_{k=1}^{j-1}(1-\sigma_k),
  \label{eq:uncertainty propagation}
\end{equation}
where $\hat{\mathbf{\beta}}(\mathbf{x}')$ denotes the 2D uncertainty of pixel $\mathbf{x}'$ and $\sigma_j$ follows the same blending weights as color rendering (\Cref{eq:splatting}).

This uncertainty propagation mechanism effectively suppresses false positives in textureless regions by grounding occlusion reasoning in geometric confidences derived from the 3D representation, enabling robust occlusion segmentation in complex outdoor scenes with dynamic lighting and transient objects.

\subsubsection{Boundary-Aware Refinement}
\label{sec:boundary-aware refinement}
Due to the lack of observations in training sets, the scene boundaries of the 3D representation often present lower rendering quality, leading to semantic discrepancies between rendered and ground-truth images and misclassifications by image models.

To address the misclassifications at scene boundaries, we introduce a boundary-aware refinement modeled as an anisotropic 2D Gaussian distribution:
\begin{equation}
    \mathcal{M}(\mathbf{x}') = e^{-\frac{1}{2}(\mathbf{x}'-\mathbf{\mu}')^\top\Sigma^{-1}(\mathbf{x}'-\mathbf{\mu}')},
  \label{eq:boundary refinement}
\end{equation}
where $\mathbf{\mu}'$ and $\Sigma$ are estimated via DINO \cite{oquab2023dinov2} features, determining the rough scope of the foreground of input images. 
This refinement modulates uncertainty values near scene peripheries, reducing sensitivity in ambiguous regions while preserving occlusion accuracy in well-observed areas.

%-------------------------------------------------------------------------
\subsection{Unified Optimization}
\label{sec:optimization}
Our training objective integrates a color loss $\mathcal{L}_\mathrm{c}$ and an uncertainty loss $\mathcal{L}_\mathrm{u}$ to jointly optimize geometry, appearance, and uncertainty reasoning. The color loss combines structural and perceptual similarity metrics through a weighted sum of DSSIM \cite{wang2004ssim} and $L_1$ losses, dynamically modulated by predicted uncertainty mask $\frac{1}{2\hat{\mathbf{\beta}}^2\mathcal{M}}$ to attenuate contributions from occluded regions:
\begin{equation}
    \mathcal{L}_c=\frac{\lambda}{2\hat{\mathbf{\beta}}^2\mathcal{M}}\mathrm{DSSIM}(\hat{\mathbf{C}},\mathbf{C}) + \frac{1-\lambda}{2\hat{\mathbf{\beta}}^2\mathcal{M}}L_1(\hat{\mathbf{C}},\mathbf{C}),
  \label{eq:color-loss}
\end{equation}
where $\lambda$ is set to 0.2 as a hyper-parameter.
The uncertainty loss $\mathcal{L}_\mathrm{u}$ enhances occlusion sensitivity through a hybrid formulation:
\begin{equation}
    \mathcal{L}_u=\frac{\mathcal{D}(\hat{\mathbf{C}},\mathbf{C})}{2\hat{\mathbf{\beta}}^2} + \lambda_1\mathrm{log}\hat{\beta},
  \label{eq:uncertainty-loss}
\end{equation}
where $\mathcal{D}(\cdot)$ measures cosine similarity between DINO \cite{oquab2023dinov2} features of rendered and ground-truth images, and $\lambda_1=0.5$ controls the trade-off between uncertainty regulation and semantic consistency.

%-------------------------------------------------------------------------

%% file: sec/4_experiments.tex
%-------------------------------------------------------------------------
\begin{table*}
    \centering
    \caption{\textbf{Quantitative Comparison} on the Photo Tourism dataset \cite{snavely2006photo}. Highlighted values indicate the \colorbox{tabfirst}{first}, \colorbox{tabsecond}{second}, and \colorbox{tabthird}{third} scores for each metric and scene.}
    \scalebox{0.95}{
    \begin{tabular}{c|ccc|ccc|ccc|cc}
         \toprule
         & \multicolumn{3}{c|}{Brandenburg Gate} & \multicolumn{3}{c|}{Sacre Coeur} & \multicolumn{3}{c|}{Trevi Fountain} & \multicolumn{2}{c}{Mean Efficiency}\\ 
         Method & \footnotesize PSNR $\uparrow$ & \footnotesize SSIM $\uparrow$ & \footnotesize LPIPS $\downarrow$ & \footnotesize PSNR $\uparrow$ & \footnotesize SSIM $\uparrow$ & \footnotesize LPIPS $\downarrow$ & \footnotesize PSNR $\uparrow$ & \footnotesize SSIM $\uparrow$ & \footnotesize LPIPS $\downarrow$ & \footnotesize GPU hrs. $\downarrow$ & \footnotesize FPS $\uparrow$\\
         \midrule
         NeRF \cite{mildenhall2021nerf} &  18.90& 0.815& 0.231& 15.60& 0.715& 0.291& 16.14& 0.600&0.366 & $>$100 & $<$1 \\
         NeRF-W-re \cite{martin2021nerfw}  & 24.17& 0.890& 0.167& 19.20& 0.807& 0.191& 18.97& 0.698& 0.265 & $>$100 & $<$1\\
         K-Planes \cite{fridovich2023kplanes} & 25.49& 0.879& 0.224& 20.61& 0.774& 0.265& 22.67& 0.714&0.317 & \cellcolor{tabfirst} 1.58 & $<$1\\
         3DGS \cite{kerbl20233d} & 19.50 & 0.871& 0.177 & 17.44& \cellcolor{tabthird} 0.836& 0.201& 17.75 & 0.706 & 0.279 & \cellcolor{tabsecond} 1.93 & \cellcolor{tabthird} 11.6\\
         GS-W \cite{zhang2024gaussianw}  & 24.32& 0.909& 0.148& 19.57& 0.826& 0.207& 20.48& 0.734& 0.252 & 11.52 &10.0\\
         WildGauss \cite{kulhanek2024wildgaussians}  & \cellcolor{tabsecond} 27.23& \cellcolor{tabfirst} 0.926& \cellcolor{tabfirst} 0.135& \cellcolor{tabthird} 22.56& \cellcolor{tabfirst} 0.859& \cellcolor{tabsecond} 0.177& 23.52& \cellcolor{tabsecond} 0.765& \cellcolor{tabfirst} 0.228 & 18.54 & 7.5\\
         \midrule
         \textbf{Ours} & \cellcolor{tabfirst} 27.76& \cellcolor{tabsecond} 0.922& \cellcolor{tabsecond} 0.141& \cellcolor{tabfirst} 23.13& \cellcolor{tabfirst} 0.859& \cellcolor{tabfirst} 0.174& \cellcolor{tabfirst} 23.97& \cellcolor{tabfirst} 0.766& \cellcolor{tabsecond} 0.237& 6.81 & 11.5 \\
         w/o light. & 20.31 & 0.880 & 0.194 & 17.08 & 0.816 & 0.224 & 17.92 & 0.701 & 0.316 & 6.73 & \cellcolor{tabfirst} 15.0\\
         w/o uncert. & 27.12 & \cellcolor{tabthird} 0.924& \cellcolor{tabthird} 0.143& \cellcolor{tabsecond} 22.61& \cellcolor{tabsecond} 0.857& 0.180 & \cellcolor{tabsecond} 23.92 & \cellcolor{tabthird} 0.761 & 0.253 & \cellcolor{tabthird} 5.96 & \cellcolor{tabsecond} 13.9 \\
         w/o refine. & \cellcolor{tabthird} 26.79& \cellcolor{tabsecond} 0.922& \cellcolor{tabthird} 0.143& 22.33 & \cellcolor{tabsecond} 0.857& \cellcolor{tabthird} 0.178& \cellcolor{tabthird} 23.81 & \cellcolor{tabfirst} 0.766 & \cellcolor{tabthird} 0.244 & 6.54 & 11.5\\
         \bottomrule
    \end{tabular}
    }
    \label{tab:quantity}
\end{table*}
%-------------------------------------------------------------------------
\begin{figure*}
    \begin{tikzpicture}
        \node[] (image) at (0,0) {};
        \newcommand\expandImage{
            \begin{scope}[shift={(image.south west)}]
                \begin{scope}[shift={(image.south west)},x={(image.south east)},y={(image.north west)}]
                    \pgfmathsetmacro{\xtwo}{\cropx + \cropWidth}
                    \pgfmathsetmacro{\ytwo}{\cropy + \cropHeight}
                    \draw[red, thick] (\cropx,\cropy) rectangle (\xtwo,\ytwo);
                    \pgfmathsetmacro{\xtwo}{\scropx + \cropWidth}
                    \pgfmathsetmacro{\ytwo}{\scropy + \cropHeight}
                    \draw[orange, thick] (\scropx,\scropy) rectangle (\xtwo,\ytwo);
                \end{scope}
                \pgfmathsetmacro{\abscropx}{\cropx * \imageWidth}
                \pgfmathsetmacro{\abscropy}{\cropy * \imageHeight}
                \pgfmathsetmacro{\cropt}{(1 - (\cropx + \cropWidth)) * \imageWidth}
                \pgfmathsetmacro{\cropl}{(1 - (\cropy + \cropHeight)) * \imageHeight}
            
                \node[anchor=north west,inner sep=0,draw=red,line width=2pt,yshift=-0.001\textwidth] (bottom_image) at (image.south west) {
                 \includegraphics[width=0.09\textwidth,trim={{\abscropx} {\abscropy} {\cropt} {\cropl}},clip]{\imagePath}};
                 
                \pgfmathsetmacro{\abscropx}{\scropx * \imageWidth}
                \pgfmathsetmacro{\abscropy}{\scropy * \imageHeight}
                \pgfmathsetmacro{\cropt}{(1 - (\scropx + \cropWidth)) * \imageWidth}%
                \pgfmathsetmacro{\cropl}{(1 - (\scropy + \cropHeight)) * \imageHeight}
                \node[anchor=north east,inner sep=0,draw=orange,line width=2pt,yshift=-0.001\textwidth] at (image.south east) {
                 \includegraphics[width=0.09\textwidth,trim={{\abscropx} {\abscropy} {\cropt} {\cropl}},clip]{\imagePath}
                };
            \end{scope}
        }
        \newcommand\placeLabel{
            \node[anchor=north,yshift=-18pt,xshift=-14pt,rotate=90] at (image1.west) {\small \imageLabel};
        }

        \pgfmathsetmacro{\imageWidth}{1006}
        \pgfmathsetmacro{\imageHeight}{754}
        \pgfmathsetmacro{\cropWidth}{200 / \imageWidth}
        \pgfmathsetmacro{\cropHeight}{130 / \imageHeight}
        \pgfmathsetmacro{\cropx}{180 / \imageWidth}
        \pgfmathsetmacro{\cropy}{5 / \imageHeight}
        \pgfmathsetmacro{\scropx}{150 / \imageWidth}
        \pgfmathsetmacro{\scropy}{550 / \imageHeight}
        \def\imageLabel{Trevi Fountain}
        \def\imagePath{images/comparison/trevi/3dgs.png}
        \node[anchor=north west,inner sep=0,alias=image,line width=0pt] (image1) at (0,0) {\includegraphics[width=0.19\textwidth]{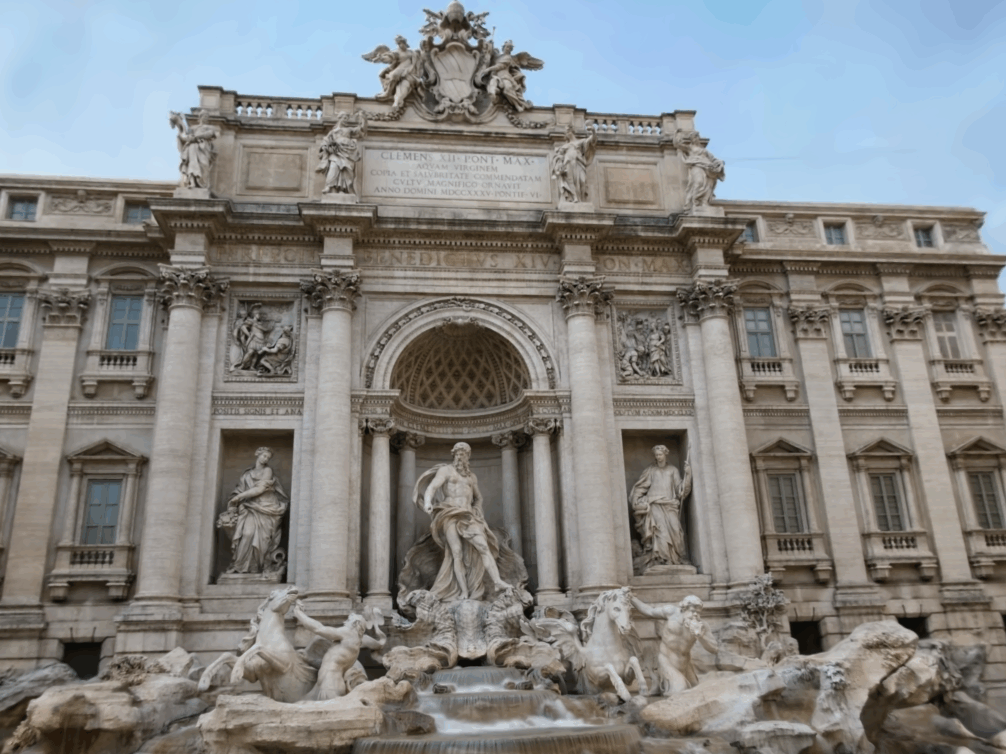}};
        \node[anchor=north, inner sep=2pt, yshift=0.023\textwidth] at (image.north) {\small 3DGS \cite{kerbl20233d}};
        \expandImage
        \node[anchor=south west] (row-marker) at (bottom_image.south west) {};
        \placeLabel
    
        \def\imagePath{images/comparison/trevi/gs-w.png}
        \node[anchor=south west,inner sep=0,alias=image,line width=0pt,xshift=0.195\textwidth] (image2) at (image1.south west) {\includegraphics[width=0.19\textwidth]{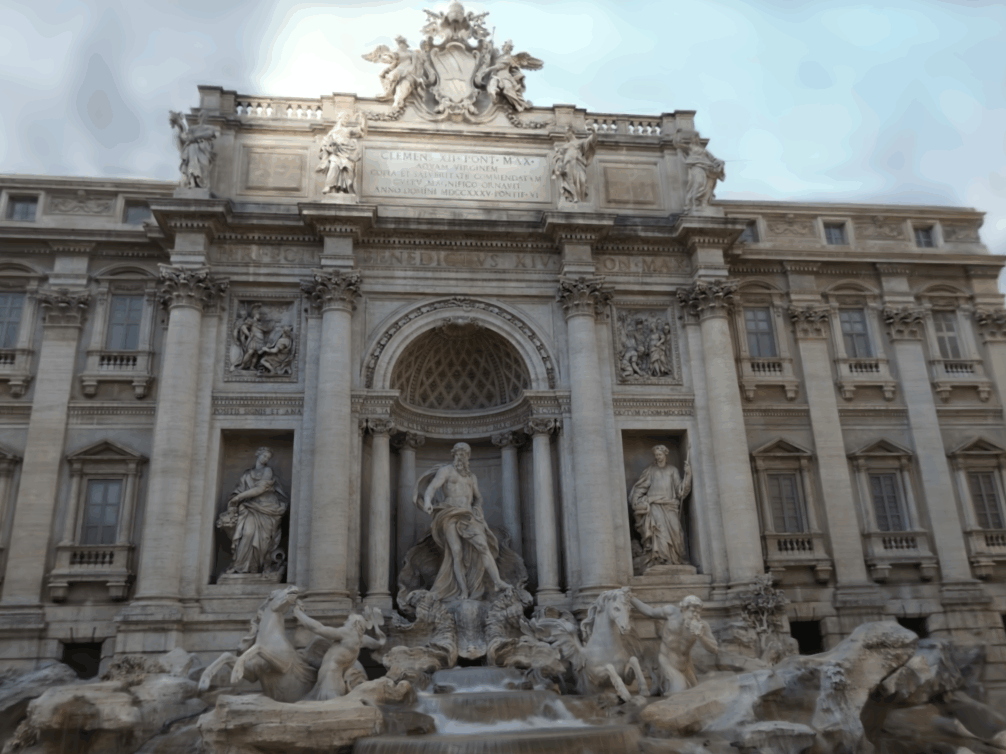}};
        \node[anchor=north, inner sep=2pt, yshift=0.023\textwidth] at (image.north) {\small GS-W \cite{zhang2024gaussianw}};
        \expandImage
        
        \def\imagePath{images/comparison/trevi/wg.png}
        \node[anchor=south west,inner sep=0,alias=image,line width=0pt,xshift=0.39\textwidth] (image3) at (image1.south west) {\includegraphics[width=0.19\textwidth]{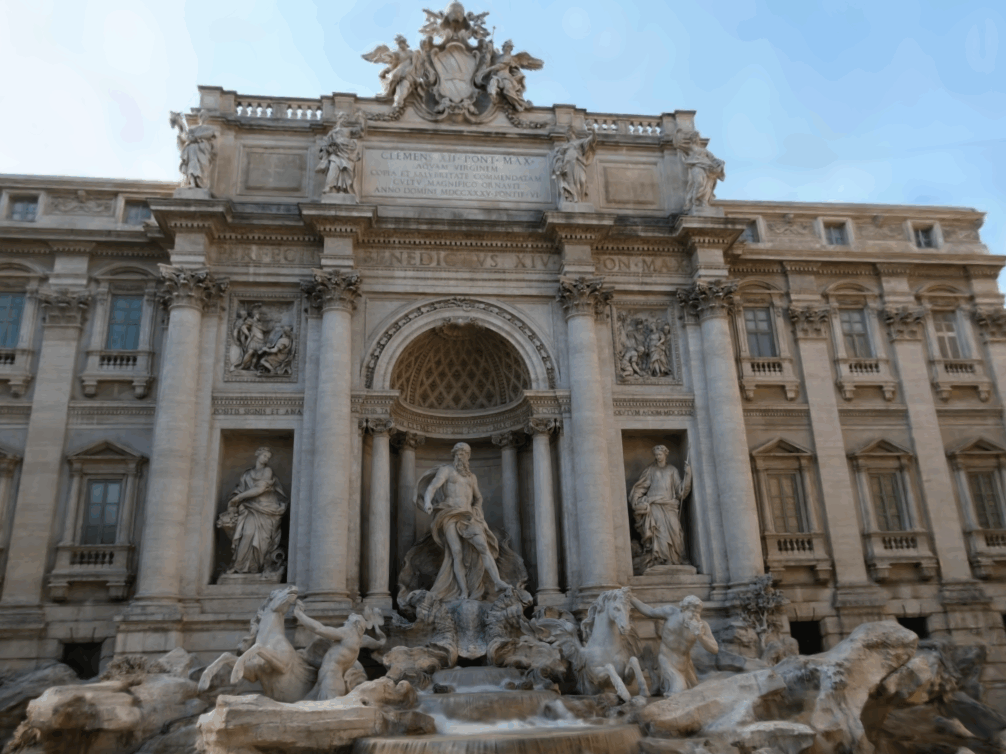}};
        \node[anchor=north, inner sep=2pt, yshift=0.023\textwidth] at (image.north) {\small WildGauss \cite{kulhanek2024wildgaussians}};
        \expandImage
    
        \def\imagePath{images/comparison/trevi/ours.png}
        \node[anchor=south west,inner sep=0,alias=image,line width=0pt,xshift=0.585\textwidth] (image4) at (image1.south west) {\includegraphics[width=0.19\textwidth]{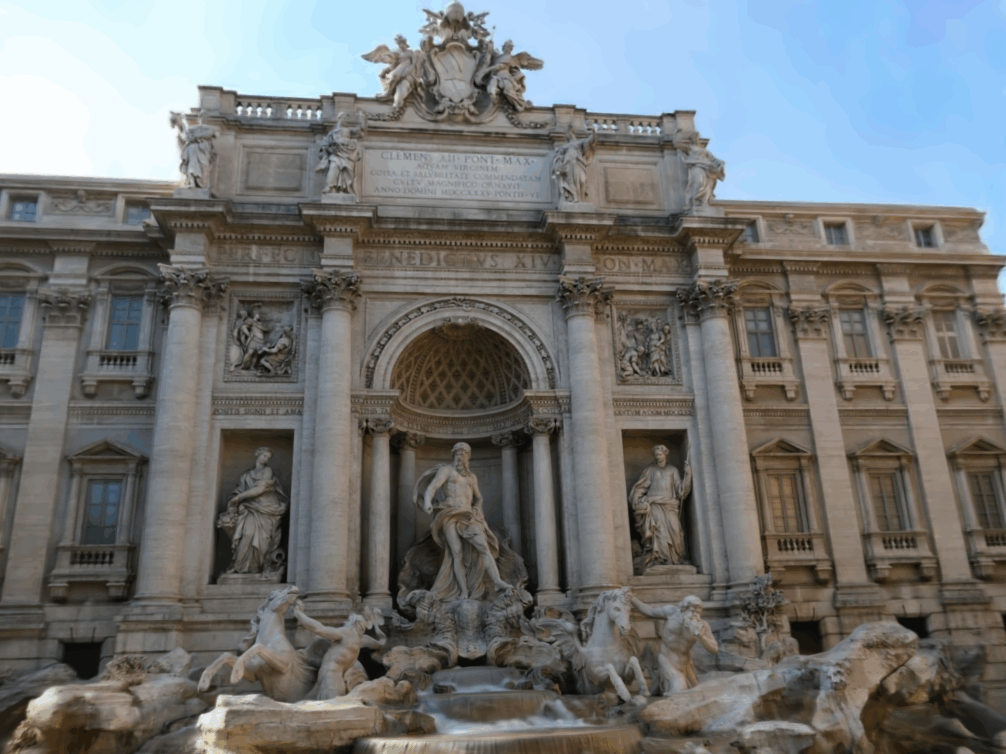}};
        \node[anchor=north, inner sep=2pt, yshift=0.023\textwidth] at (image.north) {\small  NexusSplats (\textbf{Ours})};
        \expandImage

        \def\imagePath{images/comparison/trevi/gt.png}
        \node[anchor=south west,inner sep=0,alias=image,line width=0pt,xshift=0.78\textwidth] (image4) at (image1.south west) {\includegraphics[width=0.19\textwidth]{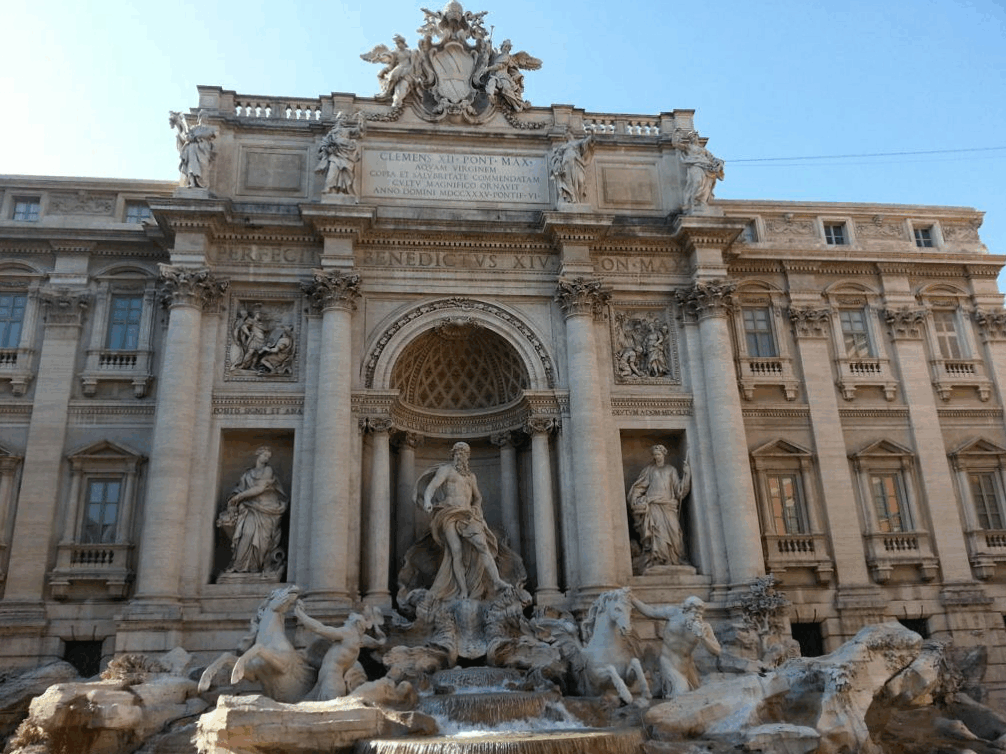}};
        \node[anchor=north, inner sep=2pt, yshift=0.023\textwidth] at (image.north) {\small Ground Truth};
        \expandImage

        \pgfmathsetmacro{\imageWidth}{1027}
        \pgfmathsetmacro{\imageHeight}{682}
        \pgfmathsetmacro{\cropWidth}{300 / \imageWidth}
        \pgfmathsetmacro{\cropHeight}{200 / \imageHeight}
        \pgfmathsetmacro{\cropx}{70 / \imageWidth}
        \pgfmathsetmacro{\cropy}{50 / \imageHeight}
        \pgfmathsetmacro{\scropx}{370 / \imageWidth}
        \pgfmathsetmacro{\scropy}{350 / \imageHeight}
        \def\imageLabel{Sacre Coeur}
        \def\imagePath{images/comparison/sacre/3dgs.png}
        \node[anchor=north west,inner sep=0,alias=image,line width=0pt,yshift=-0.005\textwidth] (image1) at (row-marker.south west) {\includegraphics[width=0.19\textwidth]{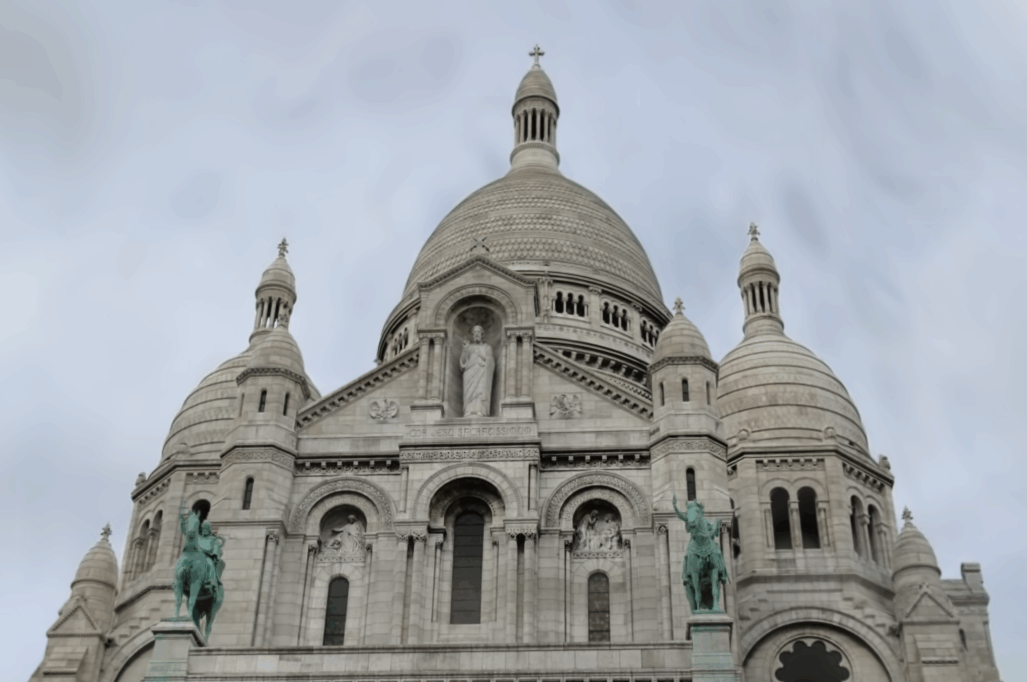}};
        \expandImage
        \node[anchor=south west] (row-marker) at (bottom_image.south west) {};
        \placeLabel
    
        \def\imagePath{images/comparison/sacre/gs-w.png}
        \node[anchor=south west,inner sep=0,alias=image,line width=0pt,xshift=0.195\textwidth] (image2) at (image1.south west) {\includegraphics[width=0.19\textwidth]{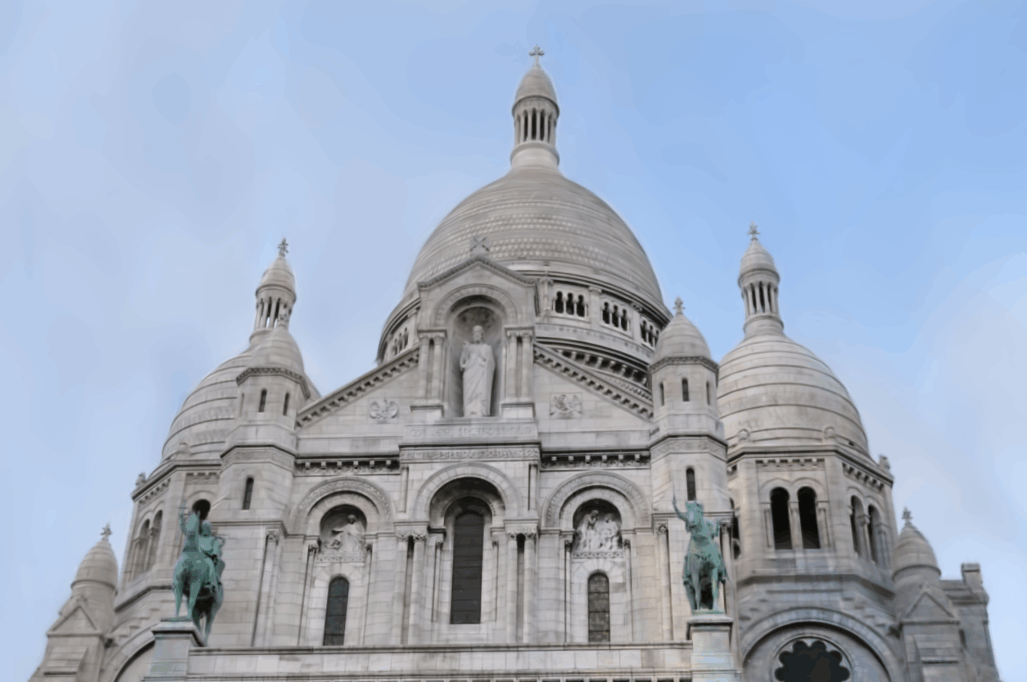}};
        \expandImage
        
        \def\imagePath{images/comparison/sacre/wg.png}
        \node[anchor=south west,inner sep=0,alias=image,line width=0pt,xshift=0.39\textwidth] (image3) at (image1.south west) {\includegraphics[width=0.19\textwidth]{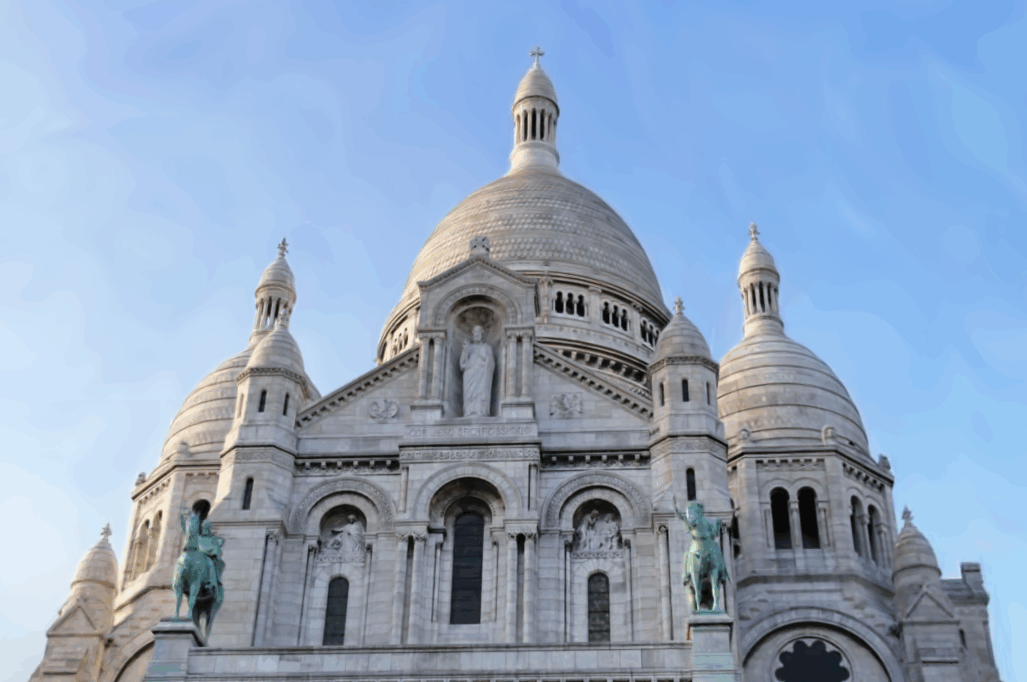}};
        \expandImage
    
        \def\imagePath{images/comparison/sacre/ours.png}
        \node[anchor=south west,inner sep=0,alias=image,line width=0pt,xshift=0.585\textwidth] (image4) at (image1.south west) {\includegraphics[width=0.19\textwidth]{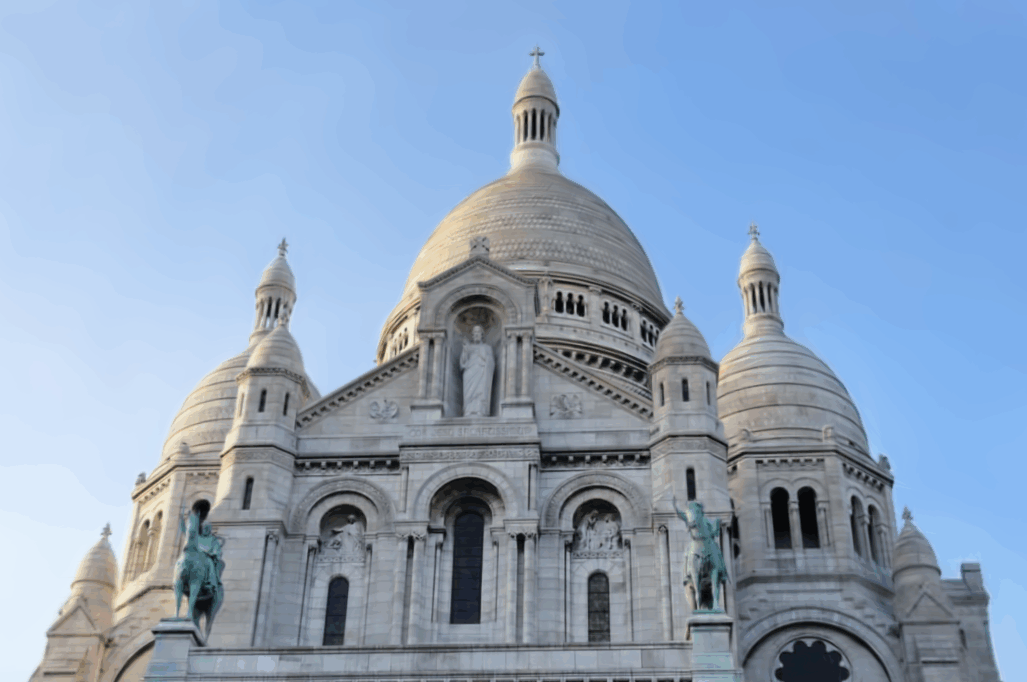}};
        \expandImage

        \def\imagePath{images/comparison/sacre/gt.png}
        \node[anchor=south west,inner sep=0,alias=image,line width=0pt,xshift=0.78\textwidth] (image4) at (image1.south west) {\includegraphics[width=0.19\textwidth]{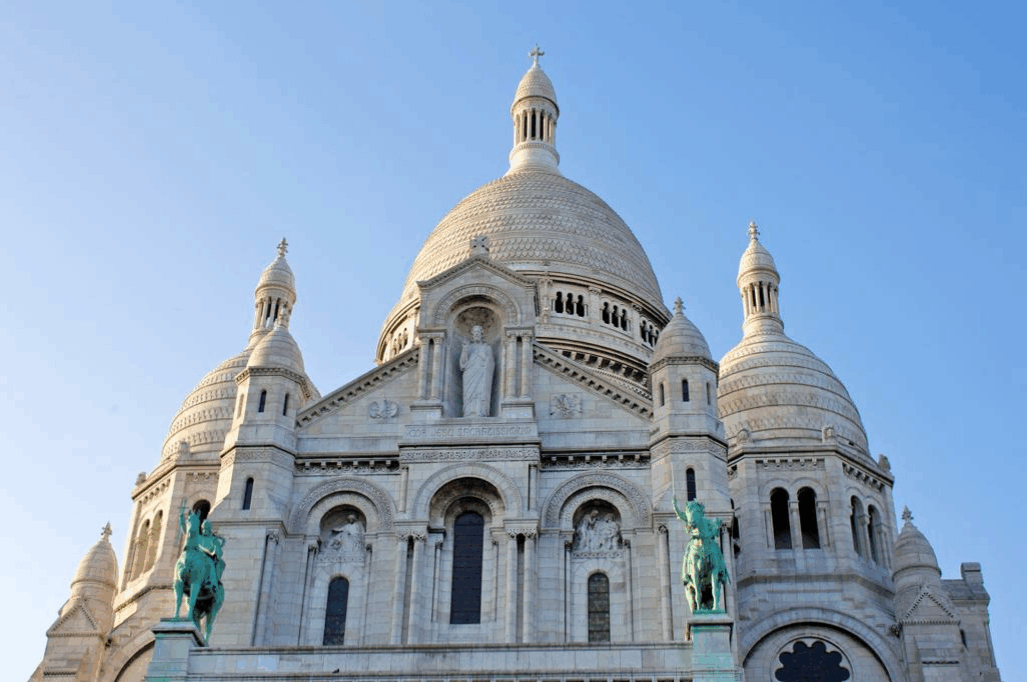}};
        \expandImage

        \pgfmathsetmacro{\imageWidth}{1049}
        \pgfmathsetmacro{\imageHeight}{777}
        \pgfmathsetmacro{\cropWidth}{300 / \imageWidth}
        \pgfmathsetmacro{\cropHeight}{200 / \imageHeight}
        \pgfmathsetmacro{\cropx}{330 / \imageWidth}
        \pgfmathsetmacro{\cropy}{30 / \imageHeight}
        \pgfmathsetmacro{\scropx}{280 / \imageWidth}
        \pgfmathsetmacro{\scropy}{520 / \imageHeight}
        \def\imageLabel{Brandenburg Gate}
        \def\imagePath{images/comparison/branden/3dgs.png}
        \node[anchor=north west,inner sep=0,alias=image,line width=0pt,yshift=-0.005\textwidth] (image1) at (row-marker.south west) {\includegraphics[width=0.19\textwidth]{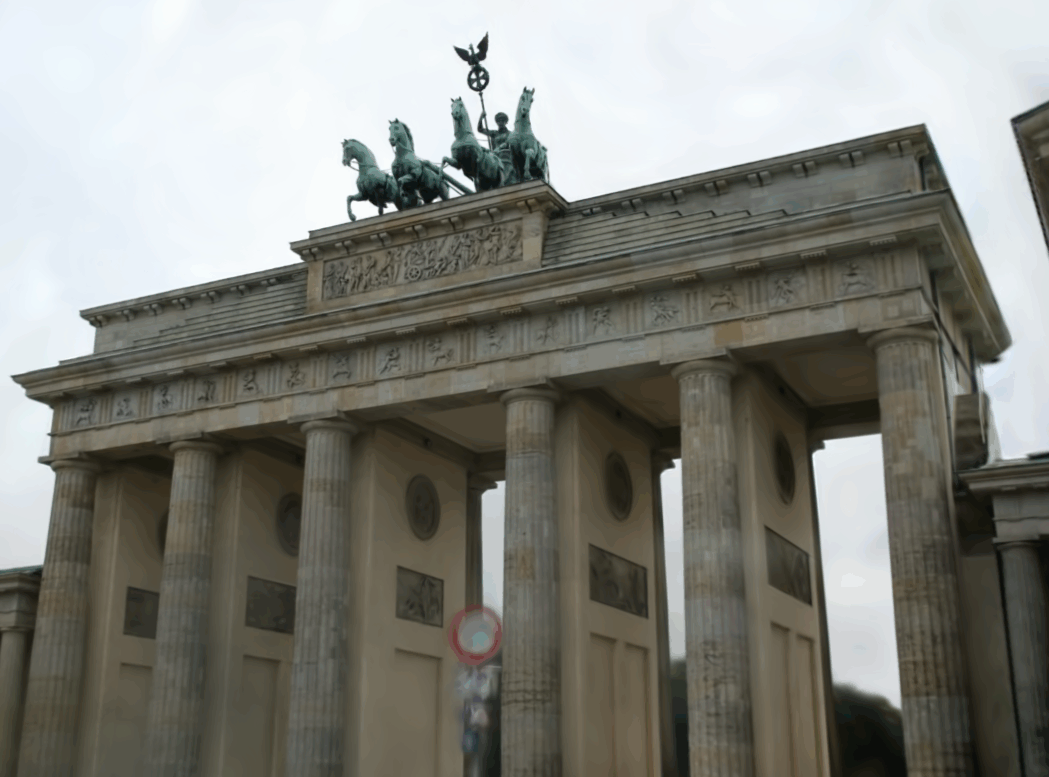}};
        \expandImage
        \placeLabel
    
        \def\imagePath{images/comparison/branden/gs-w.png}
        \node[anchor=south west,inner sep=0,alias=image,line width=0pt,xshift=0.195\textwidth] (image2) at (image1.south west) {\includegraphics[width=0.19\textwidth]{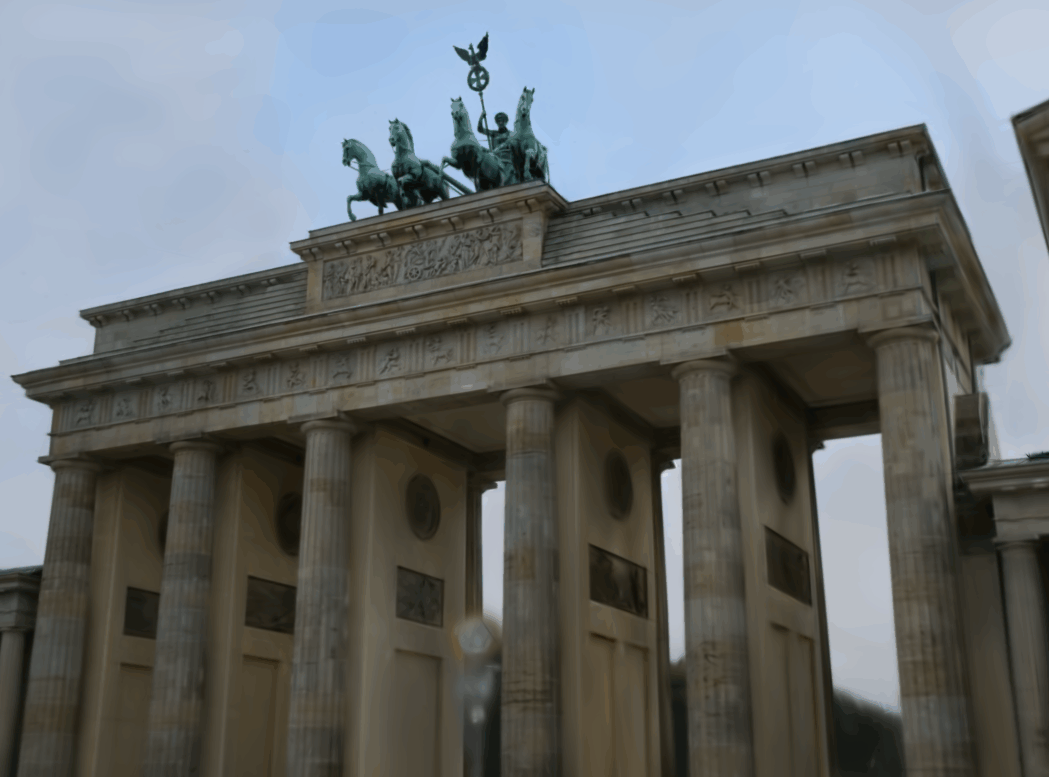}};
        \expandImage
        
        \def\imagePath{images/comparison/branden/wg.png}
        \node[anchor=south west,inner sep=0,alias=image,line width=0pt,xshift=0.39\textwidth] (image3) at (image1.south west) {\includegraphics[width=0.19\textwidth]{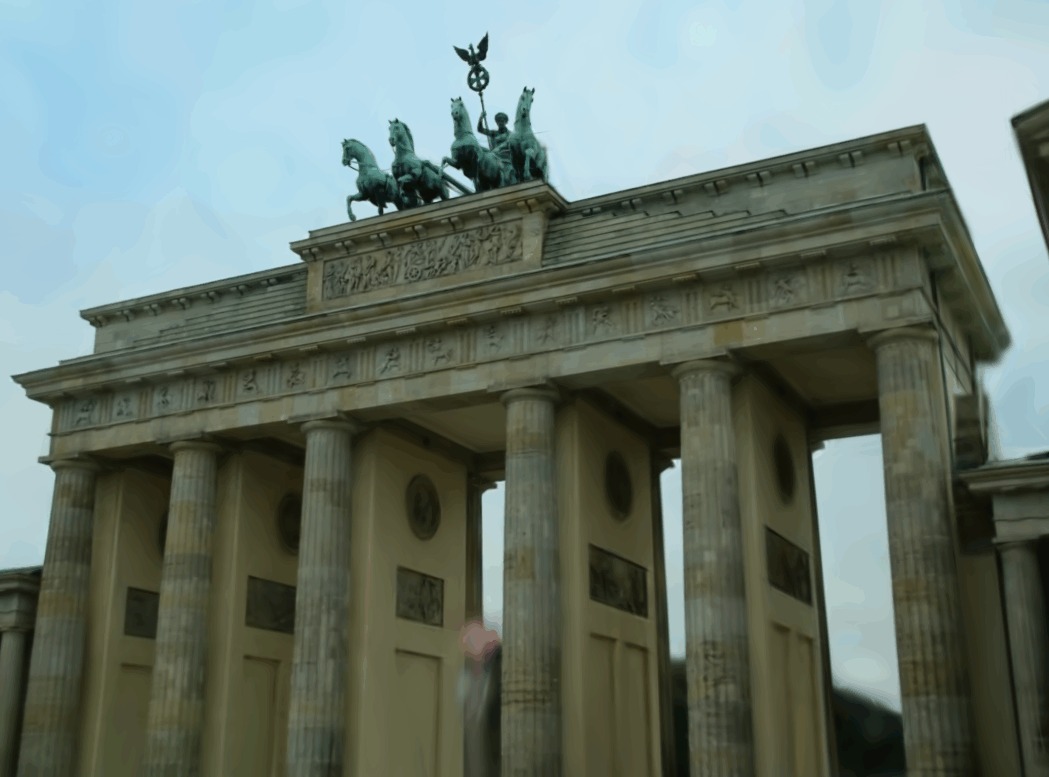}};
        \expandImage
    
        \def\imagePath{images/comparison/branden/ours.png}
        \node[anchor=south west,inner sep=0,alias=image,line width=0pt,xshift=0.585\textwidth] (image4) at (image1.south west) {\includegraphics[width=0.19\textwidth]{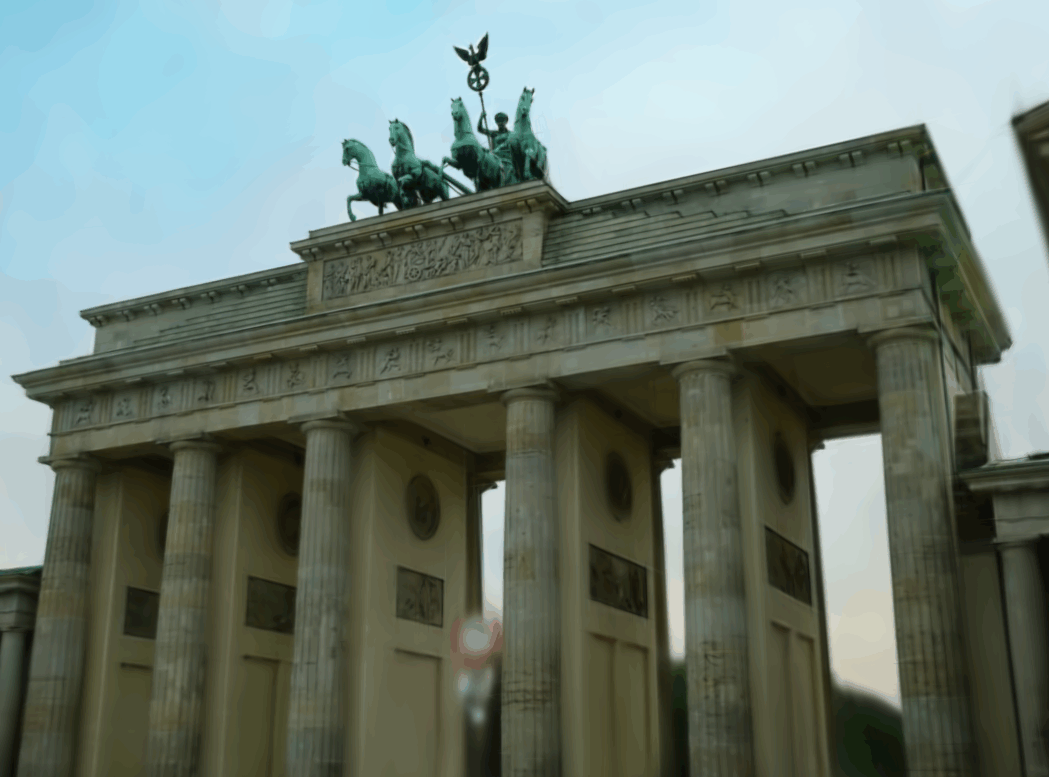}};
        \expandImage

        \def\imagePath{images/comparison/branden/gt.png}
        \node[anchor=south west,inner sep=0,alias=image,line width=0pt,xshift=0.78\textwidth] (image4) at (image1.south west) {\includegraphics[width=0.19\textwidth]{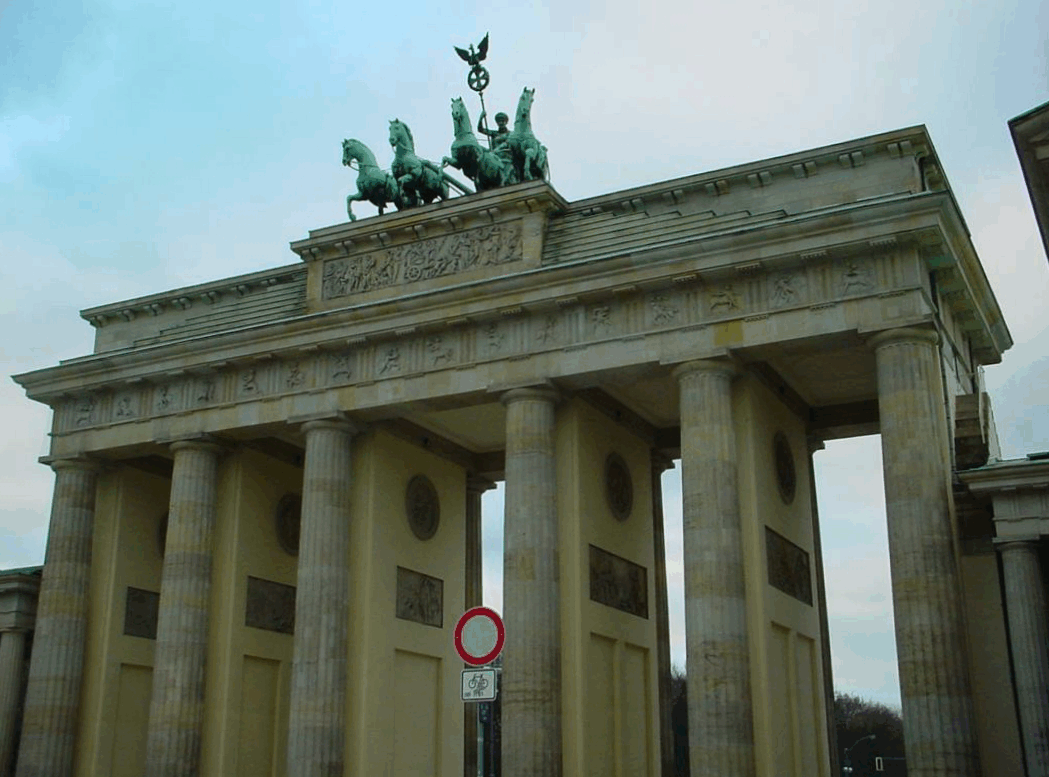}};
        \expandImage
    \end{tikzpicture}
    \vspace{-0.2in}
    \caption{\textbf{Qualitative Comparison} on the Photo Tourism dataset \cite{snavely2006photo}. For each row, different architectural scenes are shown with highlighted red and orange boxes for closer inspection. NexusSplats presents sharper details and improved color fidelity, closely matching the ground truth images, particularly in challenging areas with intricate textures.}
    \label{fig:quality}
    \vspace{-0.5em}
\end{figure*}
%-------------------------------------------------------------------------
\section{Experiments}
\subsection{Experimental Setup}
\paragraph{Datasets and Metrics.}
We evaluate NexusSplats on the Photo Tourism dataset \cite{snavely2006photo}, a standard benchmark for \textit{in-the-wild} 3D reconstruction featuring scenes with illumination changes and transient occlusions. For occlusion handling analysis, we extend evaluations to the NeRF On-the-go dataset \cite{ren2024nerfonthego}, which provides urban scenes with varying ratios of occlusions. Efficiency comparisons are conducted solely on Photo Tourism, primarily because scenes in the NeRF On-the-Go dataset lack significant lighting variations, and most baseline approaches do not incorporate light decoupling modules when reconstructing such scenes. Metrics include PSNR, SSIM \cite{wang2004ssim}, and LPIPS \cite{zhang2018unreasonable} for rendering quality, alongside training time (GPU hours) and rendering speed (FPS).

\paragraph{Baselines and Implementation.}
To ensure rigorous and reproducible evaluation, we benchmark our method against open-source methods addressing \textit{in-the-wild} novel view synthesis: NeRF variants (NeRF \cite{mildenhall2021nerf}, NeRF-W \cite{martin2021nerfw}, K-Planes \cite{fridovich2023kplanes}) and 3DGS-based approaches (3DGS \cite{kerbl20233d}, GS-W \cite{zhang2024gaussianw}, WildGauss \cite{kulhanek2024wildgaussians}). Methods without public implementations (e.g. Wild-GS \cite{xu2024wildgs}, SWAG \cite{dahmani2024swag}) are excluded to mitigate potential implementation biases and ensure consistency across evaluations. All baselines are retrained under their official configurations on a single NVIDIA A40 GPU for 200k iterations, following NeRF-W's evaluation protocol \cite{martin2021nerfw,kulhanek2024nerfbaselines} (embedding optimization on test-image left halves). Additional implementation details are provided in \Cref{sec:implementation}.
%-------------------------------------------------------------------------
\begin{figure*}
    \begin{tikzpicture}
        \node[] (image) at (0,0) {};
        \newcommand\expandImage{
            \begin{scope}[shift={(image.south west)}]
                \begin{scope}[shift={(image.south west)},x={(image.south east)},y={(image.north west)}]
                    \pgfmathsetmacro{\xtwo}{\cropx + \cropWidth}
                    \pgfmathsetmacro{\ytwo}{\cropy + \cropHeight}
                    \draw[red, thick] (\cropx,\cropy) rectangle (\xtwo,\ytwo);
                    \pgfmathsetmacro{\xtwo}{\scropx + \cropWidth}
                    \pgfmathsetmacro{\ytwo}{\scropy + \cropHeight}
                    \draw[orange, thick] (\scropx,\scropy) rectangle (\xtwo,\ytwo);
                \end{scope}
                \pgfmathsetmacro{\abscropx}{\cropx * \imageWidth}
                \pgfmathsetmacro{\abscropy}{\cropy * \imageHeight}
                \pgfmathsetmacro{\cropt}{(1 - (\cropx + \cropWidth)) * \imageWidth}
                \pgfmathsetmacro{\cropl}{(1 - (\cropy + \cropHeight)) * \imageHeight}
            
                \node[anchor=north west,inner sep=0,draw=red,line width=2pt,yshift=-0.001\textwidth] (bottom_image) at (image.south west) {
                 \includegraphics[width=0.09\textwidth,trim={{\abscropx} {\abscropy} {\cropt} {\cropl}},clip]{\imagePath}};
                 
                \pgfmathsetmacro{\abscropx}{\scropx * \imageWidth}
                \pgfmathsetmacro{\abscropy}{\scropy * \imageHeight}
                \pgfmathsetmacro{\cropt}{(1 - (\scropx + \cropWidth)) * \imageWidth}%
                \pgfmathsetmacro{\cropl}{(1 - (\scropy + \cropHeight)) * \imageHeight}
                \node[anchor=north east,inner sep=0,draw=orange,line width=2pt,yshift=-0.001\textwidth] at (image.south east) {
                 \includegraphics[width=0.09\textwidth,trim={{\abscropx} {\abscropy} {\cropt} {\cropl}},clip]{\imagePath}
                };
            \end{scope}
        }
        \newcommand\placeLabel{
            \node[anchor=north,yshift=-18pt,xshift=-14pt,rotate=90] at (image1.west) {\small \imageLabel};
        }

        \pgfmathsetmacro{\imageWidth}{504}
        \pgfmathsetmacro{\imageHeight}{378}
        \pgfmathsetmacro{\cropWidth}{200 / \imageWidth}
        \pgfmathsetmacro{\cropHeight}{130 / \imageHeight}
        \pgfmathsetmacro{\cropx}{200 / \imageWidth}
        \pgfmathsetmacro{\cropy}{20 / \imageHeight}
        \pgfmathsetmacro{\scropx}{100 / \imageWidth}
        \pgfmathsetmacro{\scropy}{180 / \imageHeight}
        \def\imageLabel{Fountain (low occl.)}
        \def\imagePath{images/on-the-go/fountain/3dgs.png}
        \node[anchor=north west,inner sep=0,alias=image,line width=0pt] (image1) at (0,0) {\includegraphics[width=0.19\linewidth]{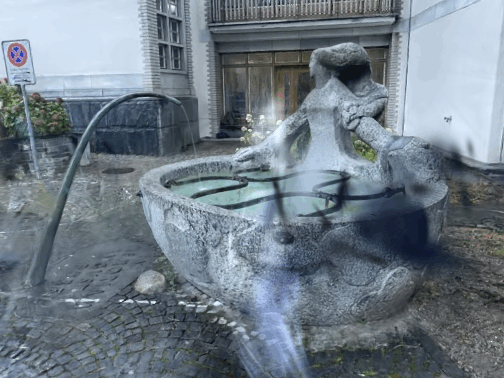}};
        \expandImage
        \node[anchor=north, inner sep=2pt,xshift=0pt,yshift=12pt] at (image.north) {\small 3DGS \cite{kerbl20233d}};
        \node[anchor=south west] (row-marker) at (bottom_image.south west) {};
        \placeLabel

        \def\imagePath{images/on-the-go/fountain/gsw.png}
        \node[anchor=south west,inner sep=0,alias=image,line width=0pt,xshift=0.195\linewidth] (image2) at (image1.south west) {\includegraphics[width=0.19\linewidth]{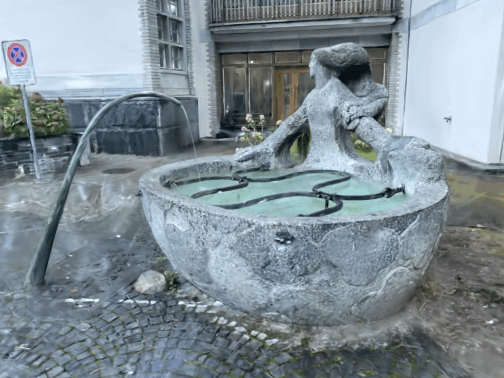}};
        \node[anchor=north, inner sep=2pt,xshift=0pt,yshift=12pt] at (image.north) {\small GS-W \cite{zhang2024gaussianw}};
        \expandImage

        \def\imagePath{images/on-the-go/fountain/wg.png}
        \node[anchor=south west,inner sep=0,alias=image,line width=0pt,xshift=0.39\linewidth] (image3) at (image1.south west) {\includegraphics[width=0.19\linewidth]{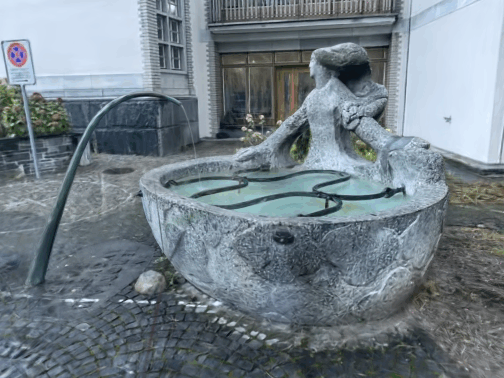}};
        \node[anchor=north, inner sep=2pt,xshift=0pt,yshift=12pt] at (image.north) {\small WildGauss \cite{kulhanek2024wildgaussians}};
        \expandImage

        \def\imagePath{images/on-the-go/fountain/ours.png}
        \node[anchor=south west,inner sep=0,alias=image,line width=0pt,xshift=0.585\linewidth] (image4) at (image1.south west) {\includegraphics[width=0.19\linewidth]{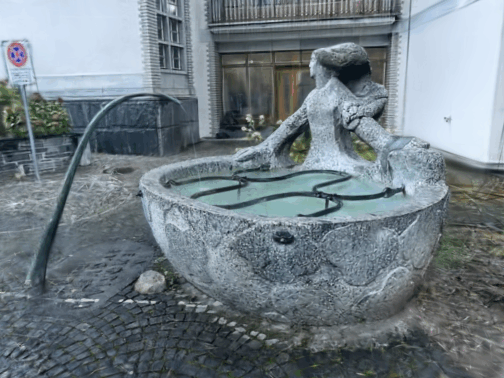}};
        \node[anchor=north, inner sep=2pt,xshift=0pt,yshift=12pt] at (image.north) {\small NexusSplats (\textbf{Ours})};
        \expandImage

        \def\imagePath{images/on-the-go/fountain/gt.png}
        \node[anchor=south west,inner sep=0,alias=image,line width=0pt,xshift=0.78\linewidth] (image5) at (image1.south west) {\includegraphics[width=0.19\linewidth]{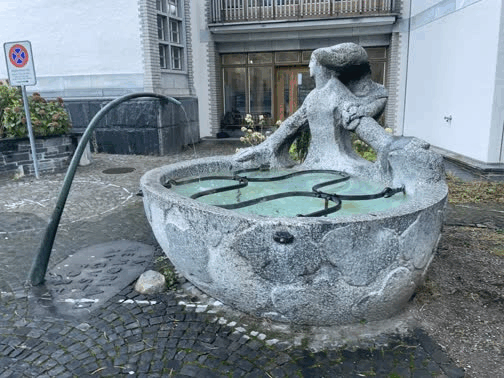}};
        \node[anchor=north, inner sep=2pt,xshift=0pt,yshift=12pt] at (image.north) {\small Ground Truth};
        \expandImage
        
        \pgfmathsetmacro{\cropx}{10 / \imageWidth}
        \pgfmathsetmacro{\cropy}{100 / \imageHeight}
        \pgfmathsetmacro{\scropx}{290 / \imageWidth}
        \pgfmathsetmacro{\scropy}{130 / \imageHeight}
        \def\imageLabel{Corner (medium occl.)}
        \def\imagePath{images/on-the-go/corner/3dgs.png}
        \node[anchor=north west,inner sep=0,alias=image,line width=0pt,yshift=-0.005\linewidth] (image1) at (row-marker.south west) {\includegraphics[width=0.19\linewidth]{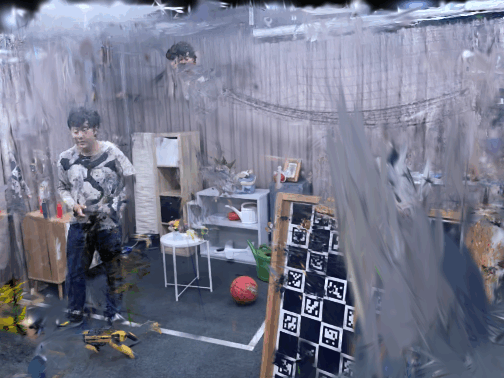}};
        \expandImage
        \node[anchor=south west] (row-marker) at (bottom_image.south west) {};
        \placeLabel

        \def\imagePath{images/on-the-go/corner/gsw.png}
        \node[anchor=south west,inner sep=0,alias=image,line width=0pt,xshift=0.195\linewidth] (image2) at (image1.south west) {\includegraphics[width=0.19\linewidth]{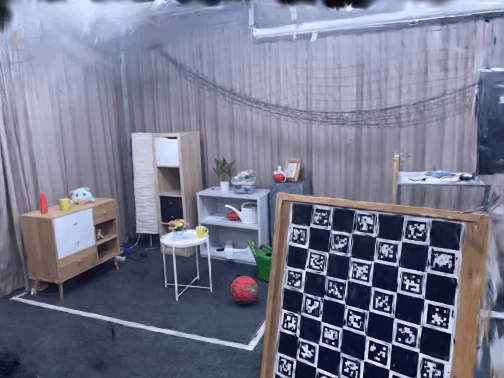}};
        \expandImage

        \def\imagePath{images/on-the-go/corner/wg.png}
        \node[anchor=south west,inner sep=0,alias=image,line width=0pt,xshift=0.39\linewidth] (image3) at (image1.south west) {\includegraphics[width=0.19\linewidth]{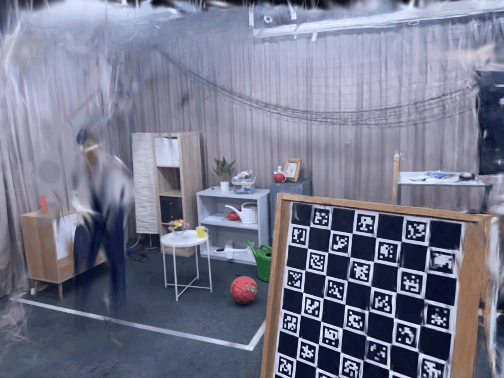}};
        \expandImage

        \def\imagePath{images/on-the-go/corner/ours.png}
        \node[anchor=south west,inner sep=0,alias=image,line width=0pt,xshift=0.585\linewidth] (image4) at (image1.south west) {\includegraphics[width=0.19\linewidth]{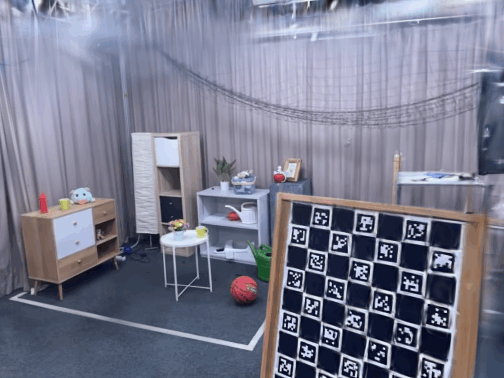}};
        \expandImage

        \def\imagePath{images/on-the-go/corner/gt.png}
        \node[anchor=south west,inner sep=0,alias=image,line width=0pt,xshift=0.78\linewidth] (image5) at (image1.south west) {\includegraphics[width=0.19\linewidth]{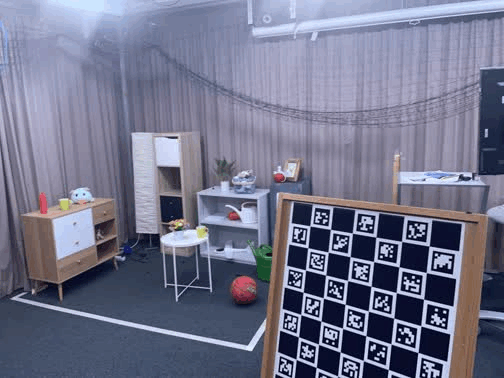}};
        \expandImage

        \pgfmathsetmacro{\cropx}{30 / \imageWidth}
        \pgfmathsetmacro{\cropy}{180 / \imageHeight}
        \pgfmathsetmacro{\scropx}{280 / \imageWidth}
        \pgfmathsetmacro{\scropy}{200 / \imageHeight}
        \def\imageLabel{Spot (high occl.)}
        \def\imagePath{images/on-the-go/spot/3dgs.png}
        \node[anchor=north west,inner sep=0,alias=image,line width=0pt,yshift=-0.005\linewidth] (image1) at (row-marker.south west) {\includegraphics[width=0.19\linewidth]{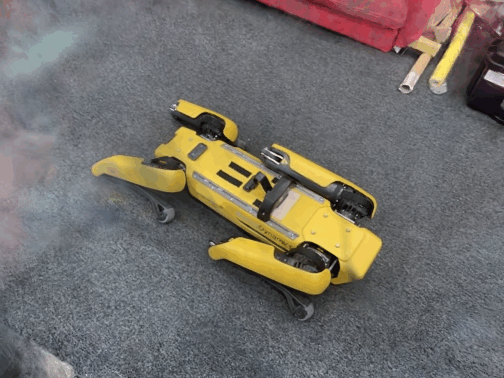}};
        \expandImage
        \placeLabel

        \def\imagePath{images/on-the-go/spot/gsw.png}
        \node[anchor=south west,inner sep=0,alias=image,line width=0pt,xshift=0.195\linewidth] (image2) at (image1.south west) {\includegraphics[width=0.19\linewidth]{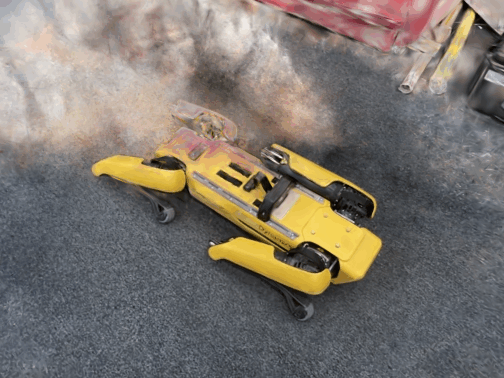}};
        \expandImage

        \def\imagePath{images/on-the-go/spot/wg.png}
        \node[anchor=south west,inner sep=0,alias=image,line width=0pt,xshift=0.39\linewidth] (image3) at (image1.south west) {\includegraphics[width=0.19\linewidth]{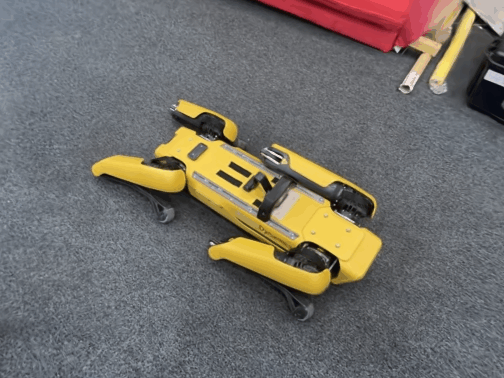}};
        \expandImage

        \def\imagePath{images/on-the-go/spot/ours.png}
        \node[anchor=south west,inner sep=0,alias=image,line width=0pt,xshift=0.585\linewidth] (image4) at (image1.south west) {\includegraphics[width=0.19\linewidth]{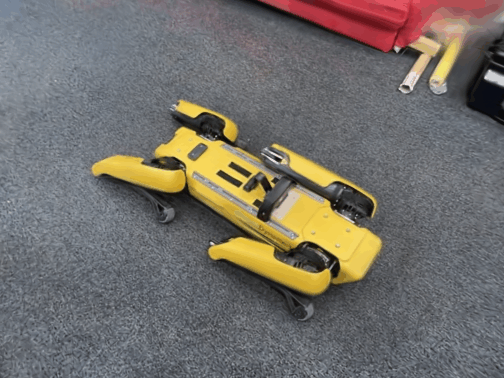}};
        \expandImage

        \def\imagePath{images/on-the-go/spot/gt.png}
        \node[anchor=south west,inner sep=0,alias=image,line width=0pt,xshift=0.78\linewidth] (image5) at (image1.south west) {\includegraphics[width=0.19\linewidth]{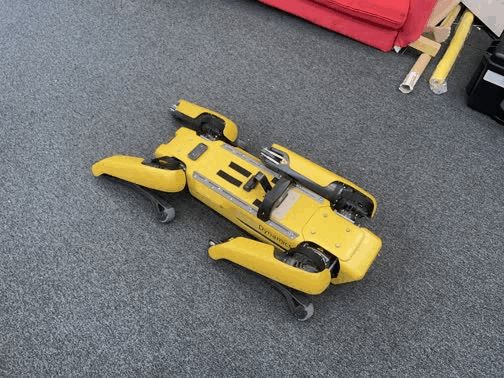}};
        \expandImage
    \end{tikzpicture}
    \vspace{-0.2in}
    \caption{\textbf{Qualitative Comparison} on the NerRF On-the-go dataset \cite{ren2024nerfonthego}. We present the results of three scenes with different ratios of occlusions. Our method removes all occlusions and shows the best view synthesis results.}
    \vspace{-0.5em}
    \label{fig:quality1}
\end{figure*}
%-------------------------------------------------------------------------
\begin{table}
    \centering
    \caption{\textbf{Efficiency Comparison} across three Photo Tourism scenes \cite{snavely2006photo}. Metrics include the number of appearance embeddings (AE), total parameters (TP), training time (TT) in GPU hours, and rendering quality (PSNR). The \colorbox{tabfirst}{first}, \colorbox{tabsecond}{second}, and \colorbox{tabthird}{third} values are highlighted.}
    \renewcommand{\arraystretch}{1.1}
    \scalebox{0.86}{
    \begin{tabular}{c|cccc}
        \toprule
        Method & AE $\downarrow$ & TP $\downarrow$ & TT $\downarrow$ & PSNR $\uparrow$ \\
        \midrule
        3DGS \cite{kerbl20233d} & - & 54.3M& \cellcolor{tabfirst} 1.93 & 18.23 \\
        GS-W \cite{zhang2024gaussianw} & \cellcolor{tabsecond} 260K & \cellcolor{tabfirst} 21.6M & \cellcolor{tabthird} 11.52& \cellcolor{tabthird} 21.46\\
        WildGauss \cite{kulhanek2024wildgaussians}& \cellcolor{tabthird} 927K& \cellcolor{tabthird} 77.1M& 18.54 & \cellcolor{tabsecond} 24.44\\
        \textbf{Ours}& \cellcolor{tabfirst} 192K & \cellcolor{tabsecond} 26.7M& \cellcolor{tabsecond} 6.81 & \cellcolor{tabfirst} 24.95 \\
        \bottomrule
    \end{tabular}}
    \label{tab:efficiency}
\end{table}
%-------------------------------------------------------------------------
\subsection{Evaluations}
\paragraph{Comparisons on Photo Tourism.}
We conduct comprehensive evaluations on the Photo Tourism dataset to assess rendering quality and efficiency under real-world conditions. Quantitative comparisons in \Cref{tab:quantity} demonstrate that NexusSplats achieves state-of-the-art rendering quality while presenting competitive training and rendering speed. Moreover, the qualitative comparison in \Cref{fig:quality} further showcases that our method preserves finer-grained details in adapting to varying lighting conditions, which corresponds the closest to the ground truth.

\paragraph{Comparisons on NeRF On-the-go.}
To validate generalization capabilities, we evaluate NexusSplats on the NeRF On-the-go dataset, which features urban scenes with transient objects. The qualitative results in \Cref{fig:quality1} demonstrate that our method successfully ignores occlusions while reconstructing static scenes across different ratios of occlusions, while other baselines incorporate transient shadows into the scene geometry at different levels. Remaining qualitative and quantitative comparisons on NeRF On-the-go are provided in \Cref{sec:full results}.

\paragraph{Efficiency Comparison.}
As shown in \Cref{tab:efficiency}, NexusSplats reduces the number of appearance embeddings by 79.3\% compared to WildGauss \cite{kulhanek2024wildgaussians}, demonstrating the parameter efficiency of kernel-level centralized appearance learning. Despite using 65.4\% fewer total parameters than WildGauss, we achieve the highest PSNR. Notably, the training time is reduced by 63.2\%, proving that our hierarchical light decoupling strategy avoids the substantial computational cost of per-Gaussian processing. Although GS-W \cite{zhang2024gaussianw} attains lower total parameters, its PSNR drops by 16.3\%, revealing a trade-off between accuracy and efficiency.
These results validate that hierarchical management enabled by nexus kernels enhances computational efficiency without damaging the fidelity of reconstruction in complex \textit{in-the-wild} scenarios.

%-------------------------------------------------------------------------
\begin{figure*}
    \begin{tikzpicture}
        \def\appa{1}
        \def\appb{54}
        \def\appc{13}
        \def\appd{652}
        \def\appe{338}
        \def\appf{1025}
        \node[] (image) at (0,0) {};
        \newcommand\placeImage{
            \pgfmathsetmacro{\abscropx}{\cropx}
            \pgfmathsetmacro{\abscropy}{\cropy}
            \pgfmathsetmacro{\cropt}{\cropx}
            \pgfmathsetmacro{\cropHeight}{0.65 * (\imageWidth - 2 * \cropx)}
            \pgfmathsetmacro{\cropl}{\imageHeight - \cropy - \cropHeight}
            \includegraphics[width=0.155\textwidth,trim={{\abscropx} {\abscropy} {\cropt} {\cropl}},clip]{\imagePath}
        }
        \newcommand\placeLabel{
            \node[anchor=south,yshift=0,xshift=13pt,rotate=90] at (image1.west) {\footnotesize \imageLabel};
        }

        \pgfmathsetmacro{\imageWidth}{1015}
        \pgfmathsetmacro{\imageHeight}{735}
        \pgfmathsetmacro{\cropx}{0}
        \pgfmathsetmacro{\cropy}{20}
        \def\imageLabel{Ground Truth}
        \def\imagePath{images/appearance/gt/\appa.jpg}
        \node[anchor=north west,inner sep=0,alias=image,line width=0pt] (image1) at (0,0) {\placeImage};
        \node[anchor=south west] (row-marker) at (image1.south west) {};
        \placeLabel
        \pgfmathsetmacro{\imageWidth}{1077}
        \pgfmathsetmacro{\imageHeight}{699}
        \pgfmathsetmacro{\cropx}{0}
        \pgfmathsetmacro{\cropy}{0}
        \def\imagePath{images/appearance/gt/\appb.jpg}
        \node[anchor=south west,inner sep=0,alias=image,line width=0pt,xshift=0.16\textwidth] (image2) at (image1.south west) {\placeImage};
        \pgfmathsetmacro{\imageWidth}{1056}
        \pgfmathsetmacro{\imageHeight}{832}
        \pgfmathsetmacro{\cropx}{50}
        \pgfmathsetmacro{\cropy}{55}
        \def\imagePath{images/appearance/gt/\appc.jpg}
        \node[anchor=south west,inner sep=0,alias=image,line width=0pt,xshift=0.32\textwidth] (image3) at (image1.south west) {\placeImage};
        \pgfmathsetmacro{\imageWidth}{1059}
        \pgfmathsetmacro{\imageHeight}{695}
        \pgfmathsetmacro{\cropx}{0}
        \pgfmathsetmacro{\cropy}{0}
        \def\imagePath{images/appearance/gt/\appd.jpg}
        \node[anchor=south west,inner sep=0,alias=image,line width=0pt,xshift=0.48\textwidth] (image4) at (image1.south west) {\placeImage};
        \pgfmathsetmacro{\imageWidth}{1061}
        \pgfmathsetmacro{\imageHeight}{692}
        \pgfmathsetmacro{\cropx}{0}
        \pgfmathsetmacro{\cropy}{0}
        \def\imagePath{images/appearance/gt/\appe.jpg}
        \node[anchor=south west,inner sep=0,alias=image,line width=0pt,xshift=0.64\textwidth] (image5) at (image1.south west) {\placeImage};
        \pgfmathsetmacro{\imageWidth}{1042}
        \pgfmathsetmacro{\imageHeight}{775}
        \pgfmathsetmacro{\cropx}{0}
        \pgfmathsetmacro{\cropy}{0}
        \def\imagePath{images/appearance/gt/\appf.jpg}
        \node[anchor=south west,inner sep=0,alias=image,line width=0pt,xshift=0.8\textwidth] (image6) at (image1.south west) {\placeImage};
        
        \pgfmathsetmacro{\imageWidth}{1035}
        \pgfmathsetmacro{\imageHeight}{772}
        \pgfmathsetmacro{\cropx}{0}
        \pgfmathsetmacro{\cropy}{40}
        \def\imageLabel{Viewpoint 1}
        \def\imagePath{images/appearance/ours1/appearance-\appa.png}
        \node[anchor=north west,inner sep=0,alias=image,line width=0pt,yshift=-0.005\textwidth] (image1) at (row-marker.south west) {\placeImage};
        \node[anchor=south west] (row-marker) at (image1.south west) {};
        \placeLabel
        \def\imagePath{images/appearance/ours1/appearance-\appb.png}
        \node[anchor=south west,inner sep=0,alias=image,line width=0pt,xshift=0.16\textwidth] (image2) at (image1.south west) {\placeImage};
        \def\imagePath{images/appearance/ours1/appearance-\appc.png}
        \node[anchor=south west,inner sep=0,alias=image,line width=0pt,xshift=0.32\textwidth] (image3) at (image1.south west) {\placeImage};
        \def\imagePath{images/appearance/ours1/appearance-\appd.png}
        \node[anchor=south west,inner sep=0,alias=image,line width=0pt,xshift=0.48\textwidth] (image4) at (image1.south west) {\placeImage};
        \def\imagePath{images/appearance/ours1/appearance-\appe.png}
        \node[anchor=south west,inner sep=0,alias=image,line width=0pt,xshift=0.64\textwidth] (image5) at (image1.south west) {\placeImage};
        \def\imagePath{images/appearance/ours1/appearance-\appf.png}
        \node[anchor=south west,inner sep=0,alias=image,line width=0pt,xshift=0.8\textwidth] (image6) at (image1.south west) {\placeImage};
        
        \pgfmathsetmacro{\imageWidth}{1070}
        \pgfmathsetmacro{\imageHeight}{690}
        \pgfmathsetmacro{\cropx}{0}
        \pgfmathsetmacro{\cropy}{0}
        \def\imageLabel{Viewpoint 2}
        \def\imagePath{images/appearance/ours3/appearance-\appa.png}
        \node[anchor=north west,inner sep=0,alias=image,line width=0pt,yshift=-0.005\textwidth] (image1) at (row-marker.south west) {\placeImage};
        \node[anchor=south west] (row-marker) at (image1.south west) {};
        \placeLabel
        \def\imagePath{images/appearance/ours3/appearance-\appb.png}
        \node[anchor=south west,inner sep=0,alias=image,line width=0pt,xshift=0.16\textwidth] (image2) at (image1.south west) {\placeImage};
        \def\imagePath{images/appearance/ours3/appearance-\appc.png}
        \node[anchor=south west,inner sep=0,alias=image,line width=0pt,xshift=0.32\textwidth] (image3) at (image1.south west) {\placeImage};
        \def\imagePath{images/appearance/ours3/appearance-\appd.png}
        \node[anchor=south west,inner sep=0,alias=image,line width=0pt,xshift=0.48\textwidth] (image4) at (image1.south west) {\placeImage};
        \def\imagePath{images/appearance/ours3/appearance-\appe.png}
        \node[anchor=south west,inner sep=0,alias=image,line width=0pt,xshift=0.64\textwidth] (image5) at (image1.south west) {\placeImage};
        \def\imagePath{images/appearance/ours3/appearance-\appf.png}
        \node[anchor=south west,inner sep=0,alias=image,line width=0pt,xshift=0.8\textwidth] (image6) at (image1.south west) {\placeImage};

        \pgfmathsetmacro{\imageWidth}{1055}
        \pgfmathsetmacro{\imageHeight}{687}
        \pgfmathsetmacro{\cropx}{0}
        \pgfmathsetmacro{\cropy}{0}
        \def\imageLabel{Viewpoint 3}
        \def\imagePath{images/appearance/ours4/appearance-\appa.png}
        \node[anchor=north west,inner sep=0,alias=image,line width=0pt,yshift=-0.005\textwidth] (image1) at (row-marker.south west) {\placeImage};
        \node[anchor=south west] (row-marker) at (image1.south west) {};
        \placeLabel
        \def\imagePath{images/appearance/ours4/appearance-\appb.png}
        \node[anchor=south west,inner sep=0,alias=image,line width=0pt,xshift=0.16\textwidth] (image2) at (image1.south west) {\placeImage};
        \def\imagePath{images/appearance/ours4/appearance-\appc.png}
        \node[anchor=south west,inner sep=0,alias=image,line width=0pt,xshift=0.32\textwidth] (image3) at (image1.south west) {\placeImage};
        \def\imagePath{images/appearance/ours4/appearance-\appd.png}
        \node[anchor=south west,inner sep=0,alias=image,line width=0pt,xshift=0.48\textwidth] (image4) at (image1.south west) {\placeImage};
        \def\imagePath{images/appearance/ours4/appearance-\appe.png}
        \node[anchor=south west,inner sep=0,alias=image,line width=0pt,xshift=0.64\textwidth] (image5) at (image1.south west) {\placeImage};
        \def\imagePath{images/appearance/ours4/appearance-\appf.png}
        \node[anchor=south west,inner sep=0,alias=image,line width=0pt,xshift=0.8\textwidth] (image6) at (image1.south west) {\placeImage};
    \end{tikzpicture}
    \vspace{-0.05in}
    \caption{\textbf{Light Decoupling Visualizations} on Photo Tourism \cite{snavely2006photo}. NexusSplats maps colors from the reconstructed scene to match the target lighting conditions from ground truth images.}
    \vspace{-0.5em}
    \label{fig:appearance}
\end{figure*}
%-------------------------------------------------------------------------
\begin{figure}
    \centering
    \setlength{\tabcolsep}{1pt}
    \def\arraystretch{1}
    \begin{tabular}{cccc}
         \footnotesize Input &
         \footnotesize 2D Uncert. \cite{kulhanek2024wildgaussians} &
         \footnotesize 3D Uncert. &
         \footnotesize Refined \\
         \includegraphics[width=0.24\linewidth]{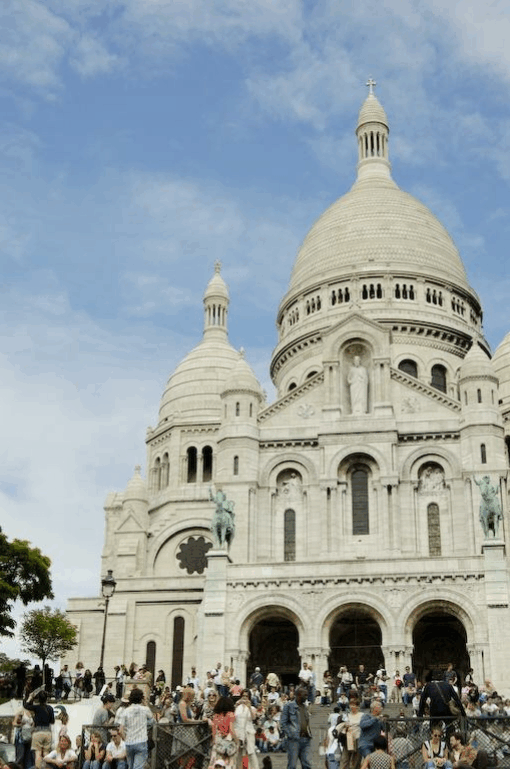} &
         \includegraphics[width=0.24\linewidth]{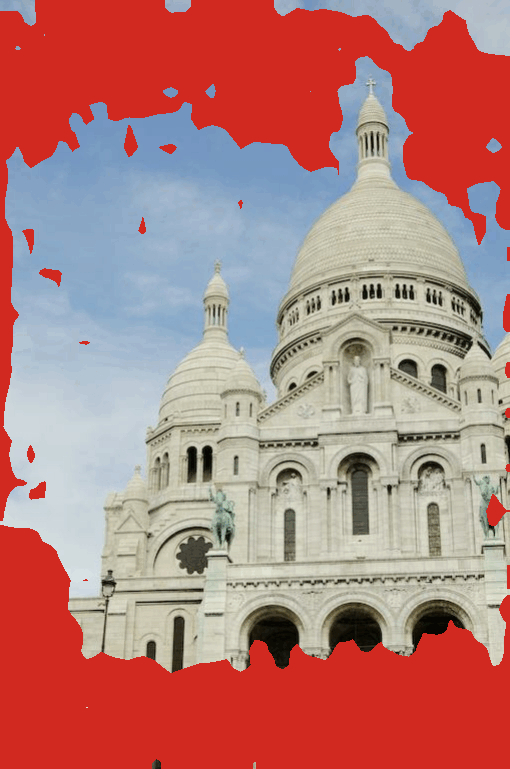} &
         \includegraphics[width=0.24\linewidth]{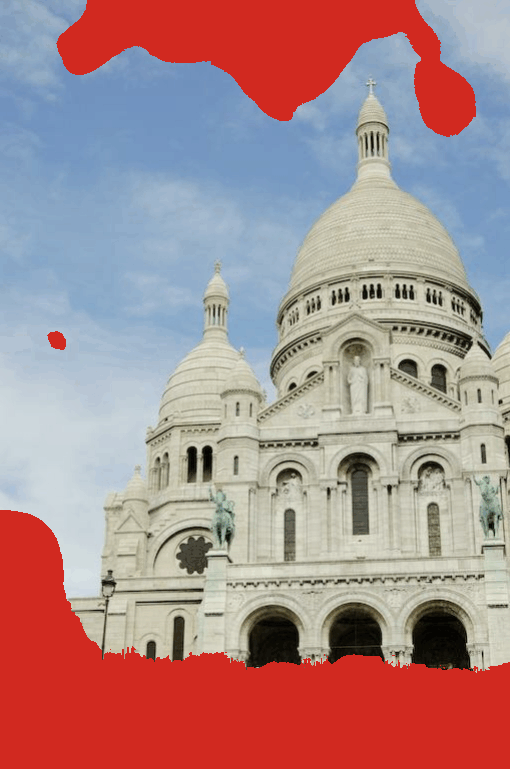} &
         \includegraphics[width=0.24\linewidth]{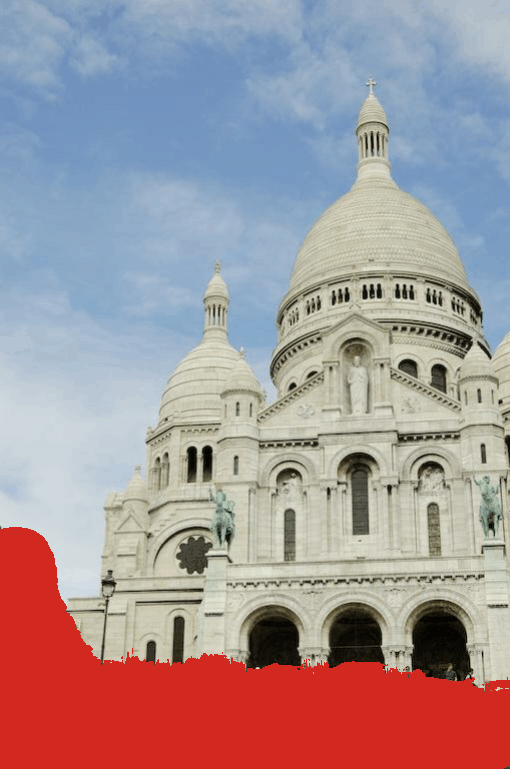} 
         \\ [-0.5mm]
         \includegraphics[width=0.24\linewidth]{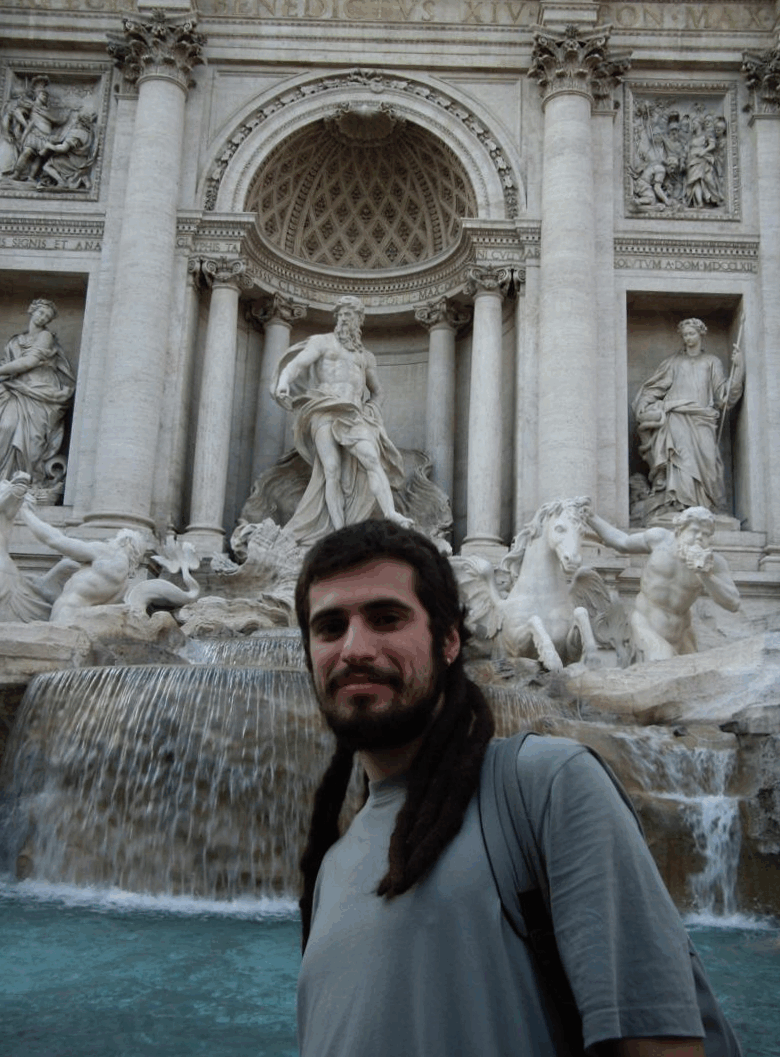} &
         \includegraphics[width=0.24\linewidth]{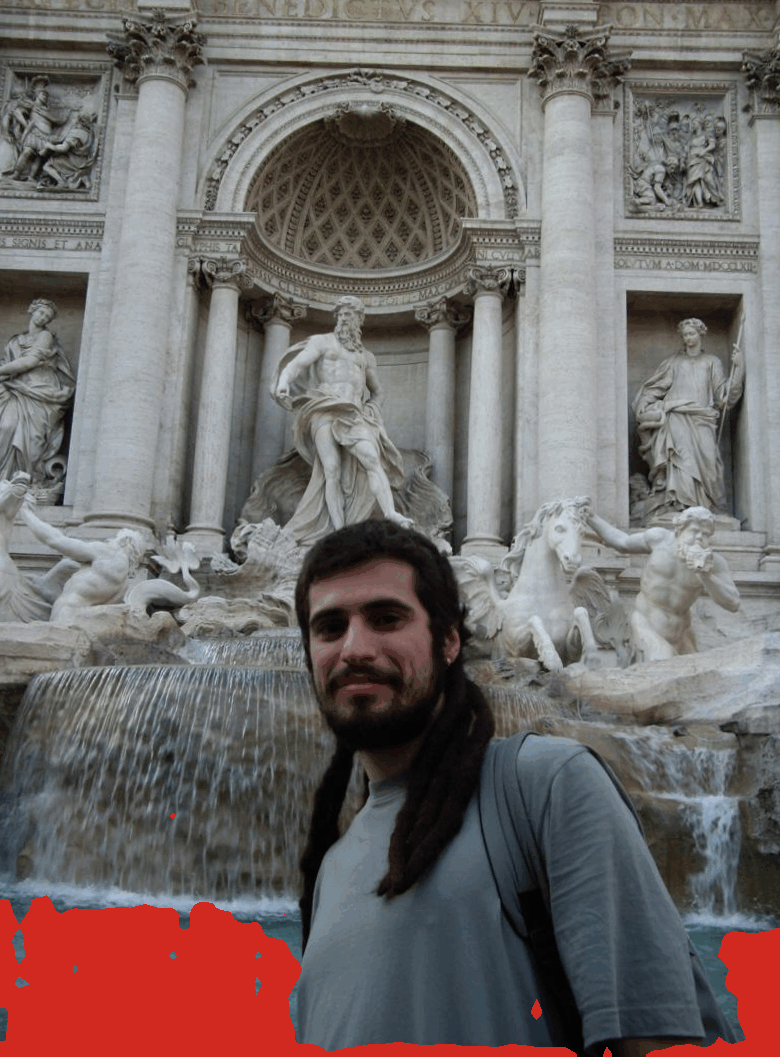} &
         \includegraphics[width=0.24\linewidth]{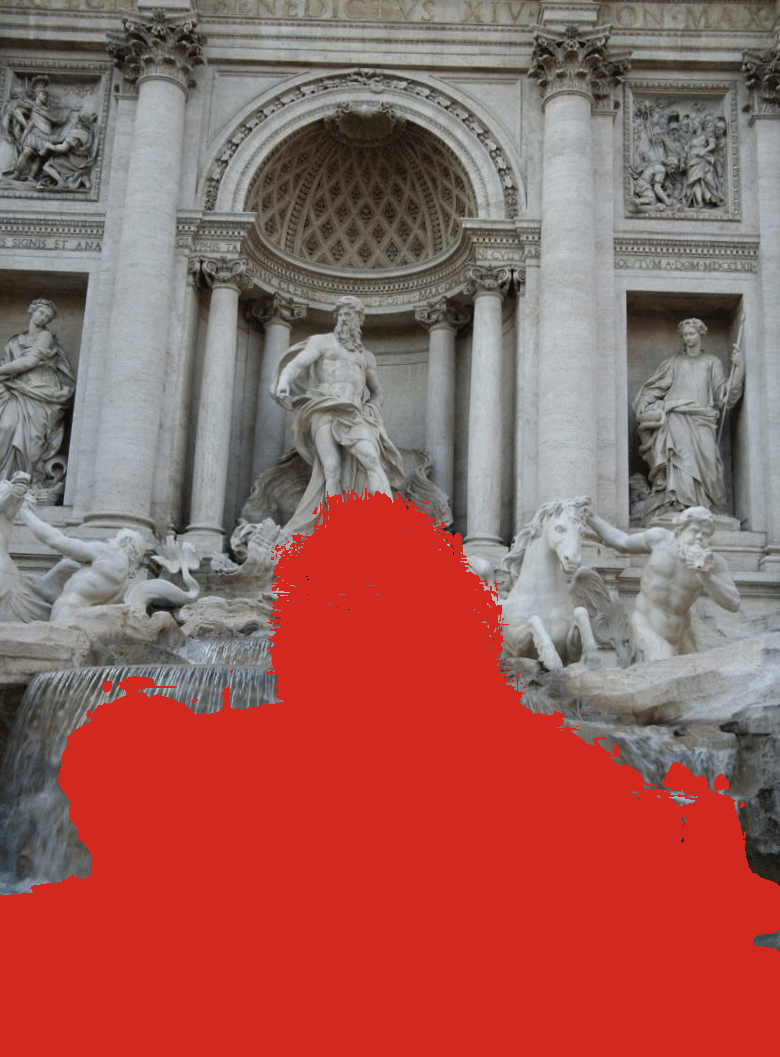} &
         \includegraphics[width=0.24\linewidth]{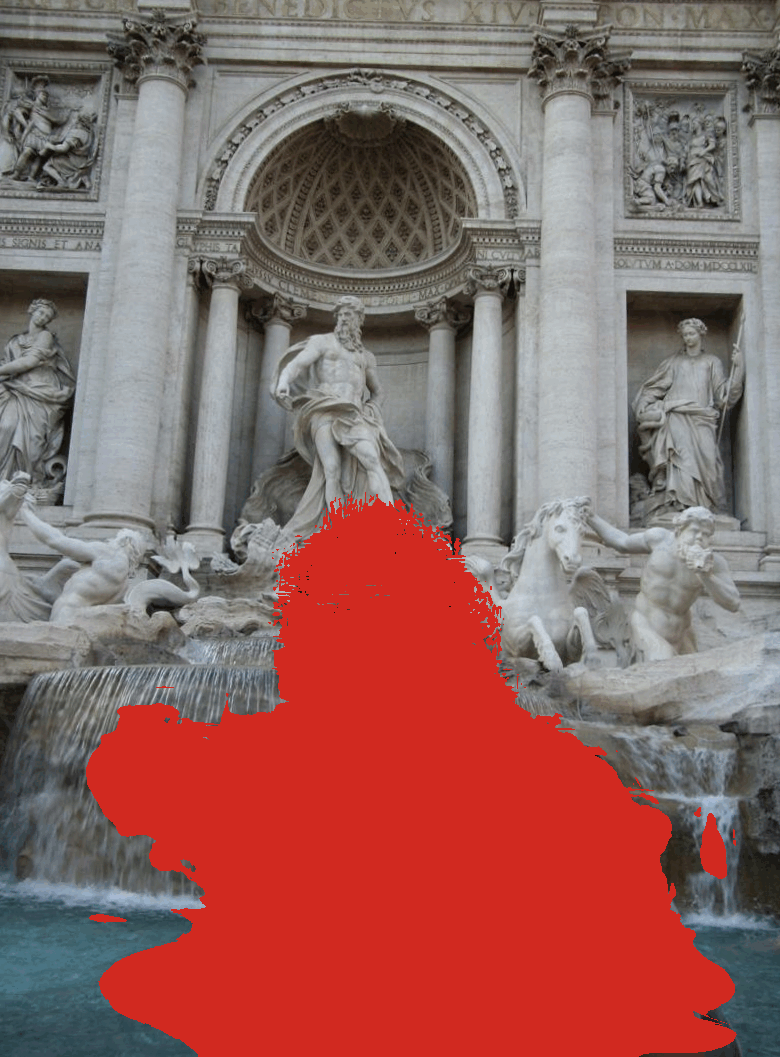} \\
    \end{tabular}
    \vspace{-0.1in}
    \caption{\textbf{Occlusion Handling Visualizations} on Photo Tourism \cite{snavely2006photo}. The red masks denote uncertainty predictions by different modelings.}
    \label{fig:uncertainty}
    \vspace{-0.5em}
\end{figure}
%-------------------------------------------------------------------------
\subsection{Ablation Study and Analysis}
To isolate the contributions of our core innovations, we systematically ablate our core modules on the Photo Tourism dataset. As shown in \Cref{tab:quantity}, disabling the \textbf{hierarchical light decoupling} module causes a severe performance drop, primarily due to unmodeled lighting variations corrupting material appearances. Removing the \textbf{structure-aware occlusion handling} module degrades reconstruction quality as transient objects pollute the static scene representation. Notably, omitting the \textbf{boundary-aware refinement} disproportionately impacts PSNR, as over-suppression of boundary regions introduces insufficient reconstruction. 
The full framework achieves optimal balance, where centralized appearance learning preserves high-frequency textures, 3D uncertainty propagation filters transient occlusions, and boundary relaxation prevents potential harm to the overall reconstruction.

\paragraph{Light Decoupling Visualizations.}
\Cref{fig:appearance} illustrates the capability of our method to decouple and adapt to varying lighting conditions. Six distinct lighting conditions from the Trevi Fountain scene, captured at different times and under various ambient lighting scenarios, serve as the target lighting conditions. NexusSplats successfully maps scene colors to these target lighting conditions across three viewpoints, demonstrating its resilience to illumination variations while maintaining color consistency and detail integrity.

\paragraph{Occlusion Handling Visualizations.}
\Cref{fig:uncertainty} showcases the improved performance in capturing occlusions of the 3D uncertainty mechanism over 2D uncertainty modeling. Furthermore, without boundary-aware refinement, our method also occasionally misidentifies clear boundary areas, such as the sky and water, as occlusions. Incorporating the refinement improves the overall performance in handling occlusions, mitigating the drawbacks of uncertainty modeling. 
Further ablations and visualizations can be found in \Cref{sec:visualization}.

%% file: sec/5_conclusion.tex
\section{Conclusion}
In this work, we introduced NexusSplats, an efficient framework for high-fidelity 3D scene reconstruction under complex lighting and occlusion conditions. By introducing hierarchical light decoupling, our method centralizes appearance learning through hierarchical Gaussian management, significantly reducing the number of parameters and accelerating training while preserving texture fidelity. The proposed structure-aware occlusion pipeline bridges 3D and 2D structural discrepancies via uncertainty propagation and boundary-aware refinement, eliminating persistent artifacts in prior works. Experimental results validate the superiority of NexusSplats in both efficiency and reconstruction quality.
Our approach marks a critical step toward practical, high-fidelity 3D reconstruction for real-world applications.

%% file: sec/X_suppl.tex
\clearpage
\setcounter{page}{1}
\maketitlesupplementary
\appendix

%-------------------------------------------------------------------------
\section*{Overview}
This appendix provides comprehensive documentation to support the claims and analyses in the main paper. It includes:
\begin{itemize}
    \item \textbf{Supplementary videos} demonstrating real-time rendering, dynamic lighting adaptation, and occlusion handling.
    \item \textbf{Implementation specifics} including network architectures, training protocols, and hyperparameters.
    \item \textbf{Further experimental results} on additional scenes and ablation studies.
    \item \textbf{Visualizations} elucidating hierarchical light decoupling and occlusion handling.
\end{itemize}
These materials collectively strengthen the validation of NexusSplats while addressing practical reproducibility concerns.

%-------------------------------------------------------------------------
\section*{Reproducibility}
To ensure full reproducibility, we have included our \textbf{source code} in the supplemental material and released our open-source implementation on GitHub, including step-by-step instructions for scene reconstruction and rendering, and pre-trained models for Photo Tourism and NeRF On-the-go datasets. 
Training is typically completed within 6–8 hours on an NVIDIA A40 GPU.

%-------------------------------------------------------------------------
\section*{Video Demo}
The supplementary videos provide dynamic, real-time demonstrations that extend beyond the static figures in the main paper. They showcase NexusSplats’ capabilities across two dimensions: multi-view renderings under varying lighting conditions and interactive side-by-side comparisons with 3DGS \cite{kerbl20233d}, GS-W \cite{zhang2024gaussianw}, and WildGauss \cite{kulhanek2024wildgaussians} in complex outdoor environments. These visualizations highlight our method’s unique strengths --- particularly its ability to harmonize lighting adaptation with geometric precision while maintaining real-time performance. We strongly encourage readers to watch these videos to fully appreciate the temporal coherence of reconstructed scenes.

%-------------------------------------------------------------------------
\section{Implementation Details}
\label{sec:implementation}

Our implementation initializes \textit{nexus kernels} from Structure-from-Motion (SfM) points \cite{schonberger2016structure}, aligning with established practices in neural scene representation \cite{kerbl20233d,xu2022point,mildenhall2021nerf}.
Each \textit{nexus kernel} manages $h=10$ neural Gaussians through learnable offsets, with a 32-dimensional local context feature vector $\mathbf{f}_u$ to encode spatial semantics, consistent with Scaffold-GS \cite{lu2024scaffold}.
Kernel pruning occurs between 1500 and 15000 iterations, removing underperforming kernels with accumulated opacity below $\tau=0.005$.

For lighting adaptation, we adopt a 32-dimensional image-specific light embedding $\varepsilon_l$ and a 30-dimensional kernel-wise appearance embedding $\varepsilon_a$.
The color mapping MLP $F_\theta$ employs two 256-unit hidden layers with ReLU activations, while a dropout rate of 0.2 optimally balances generalization and overfitting (see \Cref{tab:dropout}).

Similarly, the uncertainty propagation module uses a 32-dimensional image-specific transient embedding $\varepsilon_\tau$ and a 30-dimensional kernel uncertainty embedding $\varepsilon_\sigma$, processed by an MLP $F_\sigma$ with two 128-unit layers.

Training is conducted using PyTorch and Adam optimization, with evaluation performed via the NerfBaselines framework \cite{kulhanek2024nerfbaselines}.
We excluded methods, including HA-NeRF \cite{chen2022hallucinated}, CR-NeRF \cite{yang2023cross}, RefinedFields \cite{kassab2023refinedfields}, SWAG \cite{dahmani2024swag}, Wild-GS \cite{xu2024wildgs} and WE-GS \cite{wang2024wegs}, due to unavailable implementations or hardware constraints. We adhere strictly to dataset splits and metrics from \cite{kulhanek2024nerfbaselines} for fair comparisons.

%-------------------------------------------------------------------------
\begin{table}
    \centering
    \caption{\textbf{Overfitting Studies} on the Sacre Coeur scene, where the performance in quality drops the most, showing the results of different dropout rates in the light decoupling module. \textbf{Bold} indicates best results.}
    \renewcommand{\arraystretch}{1.1}
    \scalebox{0.95}{
\begin{tabular}{c|ccc}
         \toprule
         Dropout Rate & PSNR $\uparrow$ & SSIM $\uparrow$ & LPIPS $\downarrow$ \\
         \midrule
         0& 22.80& 0.857& 0.174\\
         0.1& 22.44& \textbf{0.859}&0.176\\
         0.2& \textbf{22.99}& \textbf{0.859}&\textbf{0.172}\\
         0.3& 22.51& 0.857&0.176\\
         0.4& 22.28& 0.853&0.179\\
         0.5& 22.32& 0.855&0.177\\
         \bottomrule
    \end{tabular}}
    \label{tab:dropout}
\end{table}
%-------------------------------------------------------------------------
\begin{table*}
    \centering
    \caption{\textbf{Quantitative Comparison} on the NeRF On-the-go dataset \cite{ren2024nerfonthego}. Highlighted values indicate the \colorbox{tabfirst}{first}, \colorbox{tabsecond}{second}, and \colorbox{tabthird}{third} scores for each metric and scene.}
    \renewcommand{\arraystretch}{1.2}
    \setlength{\tabcolsep}{0.2pt}
    \scalebox{0.88}{
    %
    \vspace{-0.2in}
    \caption{\textbf{Qualitative Comparison} on the NerRF On-the-go dataset \cite{ren2024nerfonthego}. We present the results of the remaining three scenes.}
    \vspace{-0.5em}
    \label{fig:quality2}
\end{figure*}
%-------------------------------------------------------------------------
\begin{figure*}
    %
    \vspace{-0.2in}
    \caption{\textbf{Qualitative Ablations} on the Photo Tourism dataset \cite{snavely2006photo}.}
    \label{fig:quality ablation1}
    \vspace{-0.5em}
\end{figure*}
%-------------------------------------------------------------------------
\begin{figure}
    \scalebox{0.9}{
    %
}
    % \vspace{-0.2in}
    \caption{\textbf{Qualitative Comparison} on the NerRF On-the-go dataset \cite{ren2024nerfonthego}.}
    \vspace{-0.5em}
    \label{fig:quality ablation2}
\end{figure}
%-------------------------------------------------------------------------
\begin{figure}
    \scalebox{0.9}{
    %
}
    % \vspace{-0.2in}
    \caption{\textbf{Qualitative Comparison} on the NerRF On-the-go dataset \cite{ren2024nerfonthego}.}
    \vspace{-0.5em}
    \label{fig:quality ablation3}
\end{figure}
%-------------------------------------------------------------------------
\begin{figure*}
    %
    % \vspace{-0.1in}
    \caption{\textbf{Light Decoupling Visualization} for Brandenburg Gate.}
    \vspace{-0.5em}
    \label{fig:appearance1}
\end{figure*}
%-------------------------------------------------------------------------
\begin{figure*}
    %
    % \vspace{-0.1in}
    \caption{\textbf{Light Decoupling Visualization} for Sacre Coeur.}
    \vspace{-0.5em}
    \label{fig:appearance2}
\end{figure*}
%-------------------------------------------------------------------------
\begin{figure}
    \centering
    \setlength{\tabcolsep}{1pt}
    \def\arraystretch{1}
    \begin{tabular}{cccc}
         \footnotesize Input &
         \footnotesize 2D Uncert. \cite{kulhanek2024wildgaussians} &
         \footnotesize 3D Uncert. &
         \footnotesize Refined \\
         \includegraphics[width=0.24\linewidth]{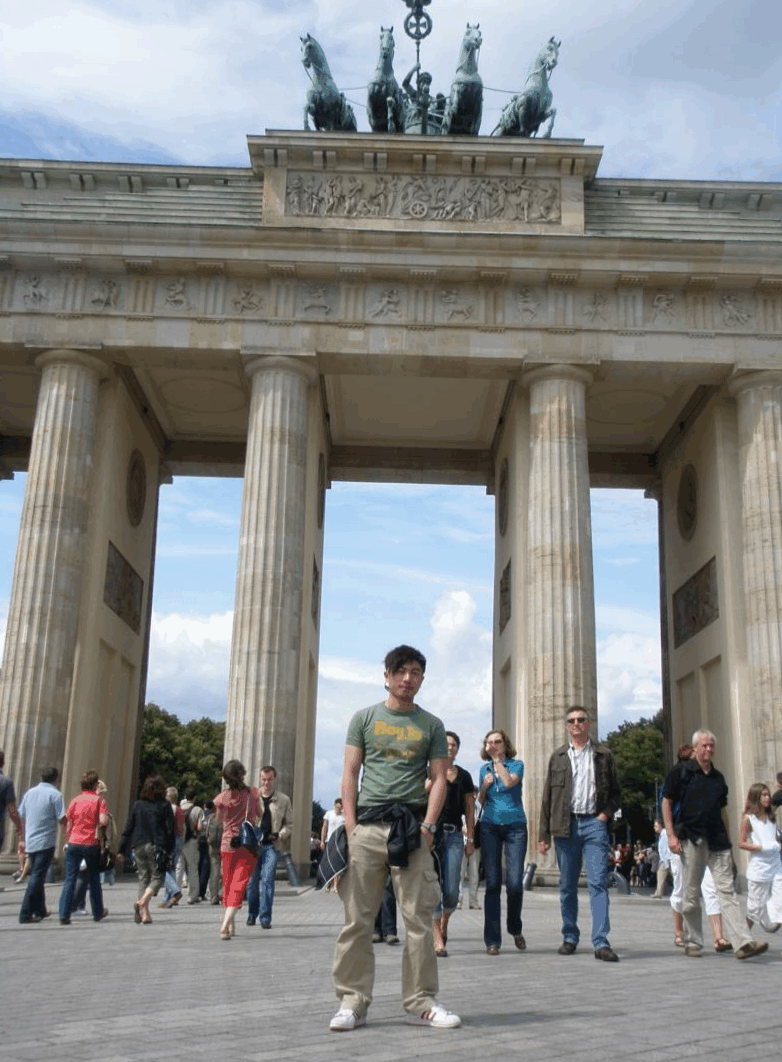} &
         \includegraphics[width=0.24\linewidth]{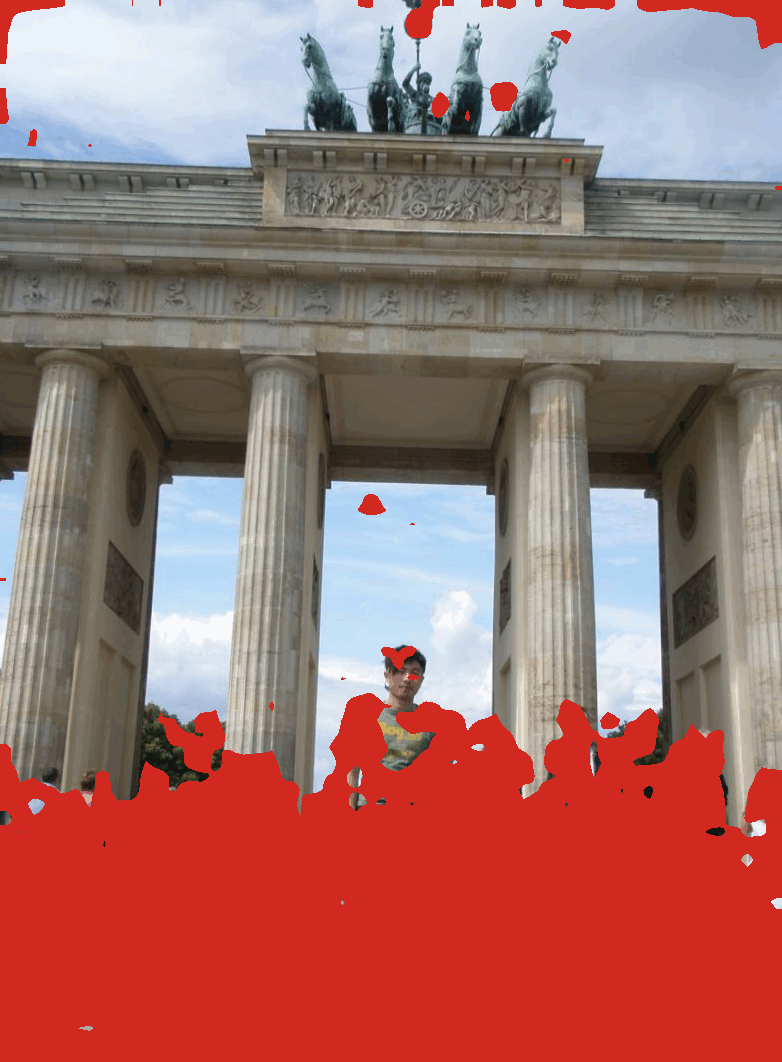} &
         \includegraphics[width=0.24\linewidth]{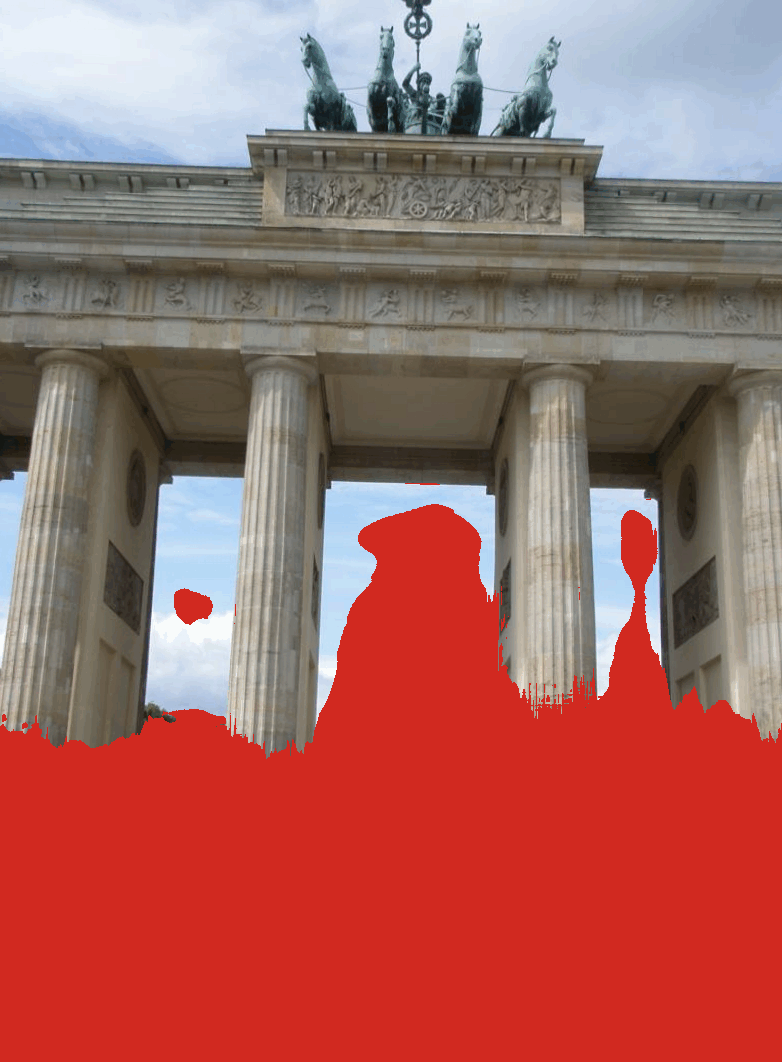} &
         \includegraphics[width=0.24\linewidth]{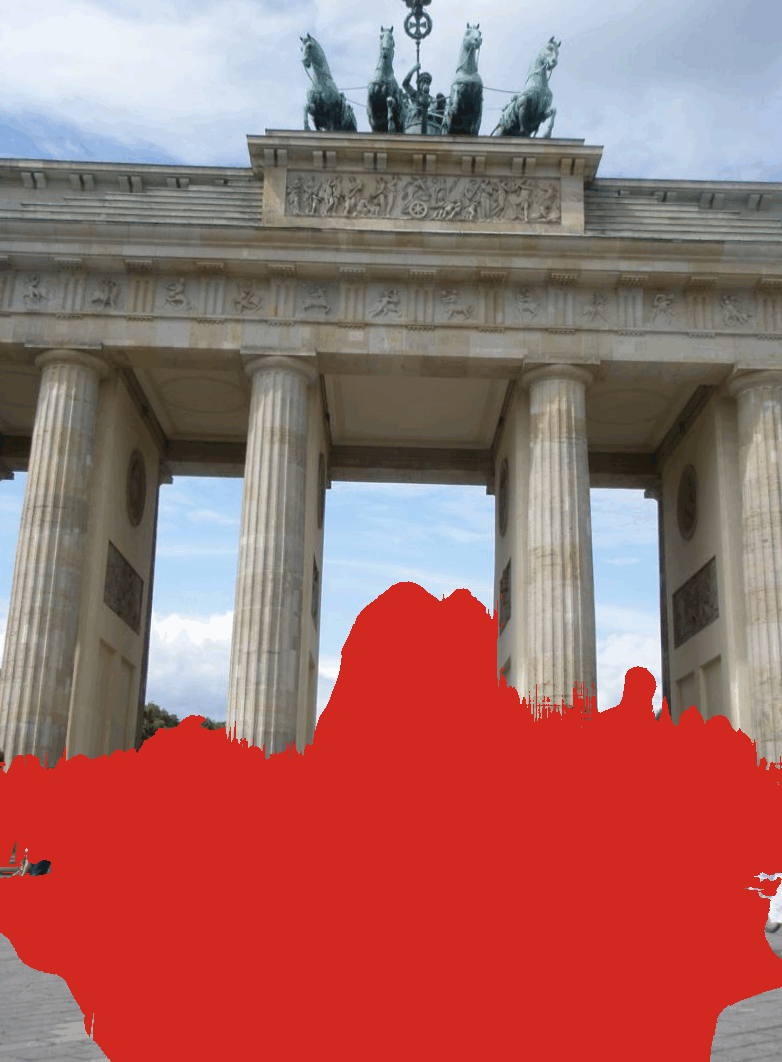} 
         \\ [-0.5mm]
         \includegraphics[width=0.24\linewidth]{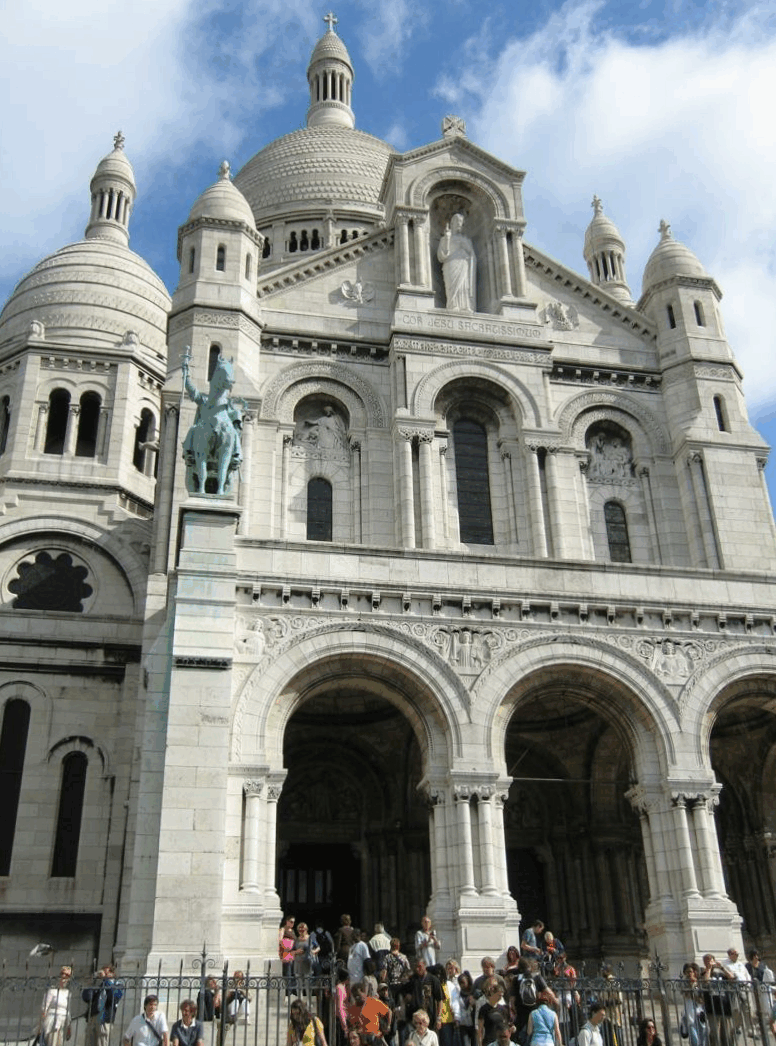} &
         \includegraphics[width=0.24\linewidth]{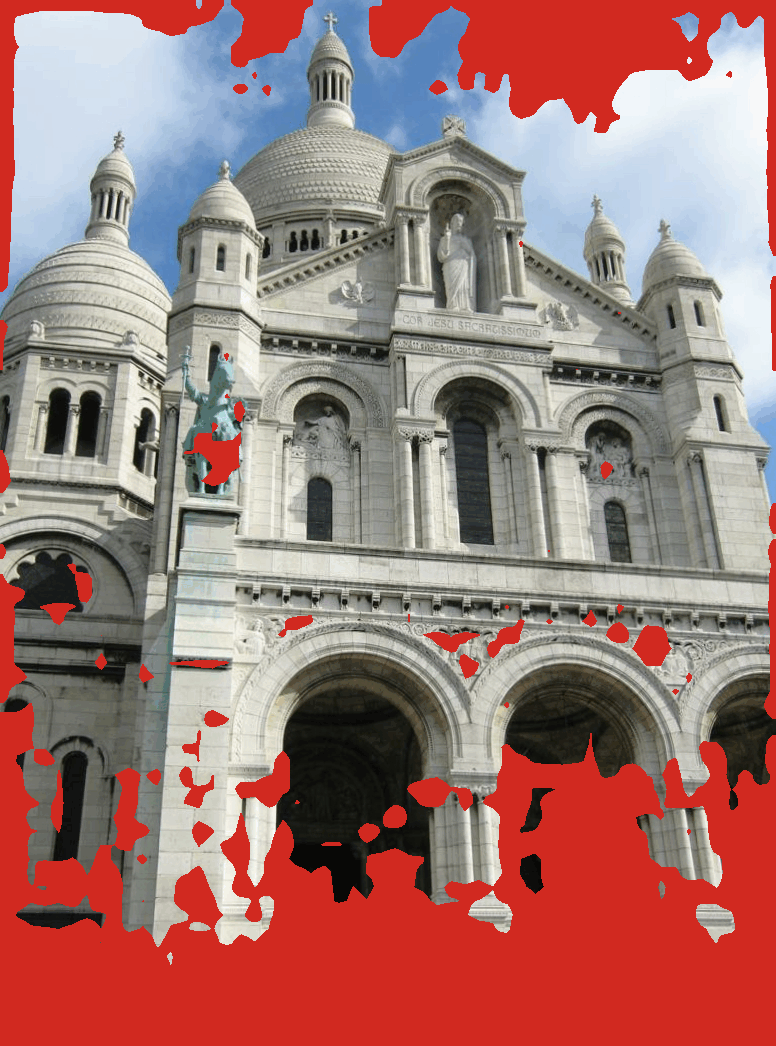} &
         \includegraphics[width=0.24\linewidth]{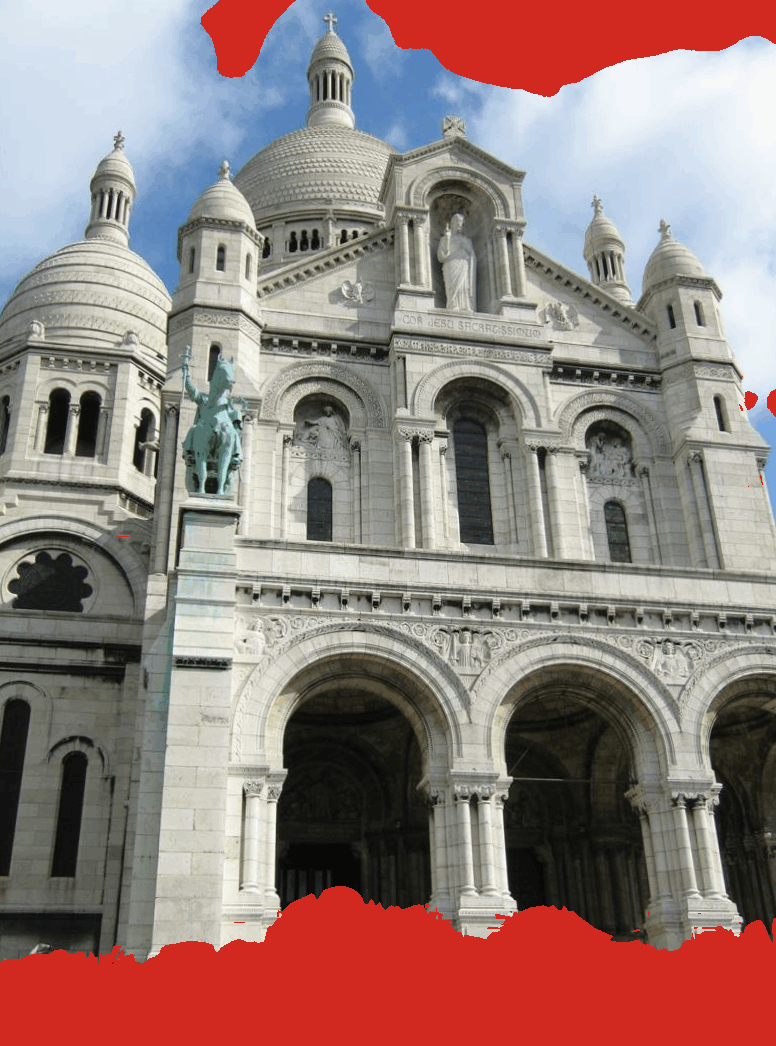} &
         \includegraphics[width=0.24\linewidth]{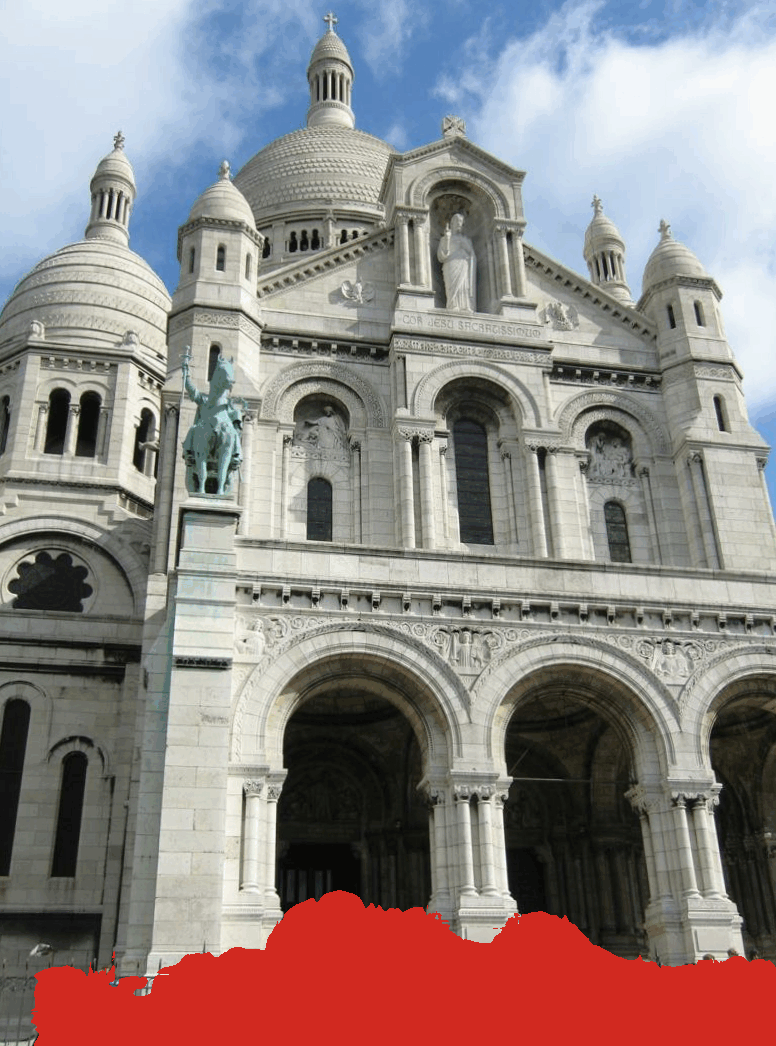} 
         \\ [-0.5mm]
         \includegraphics[width=0.24\linewidth]{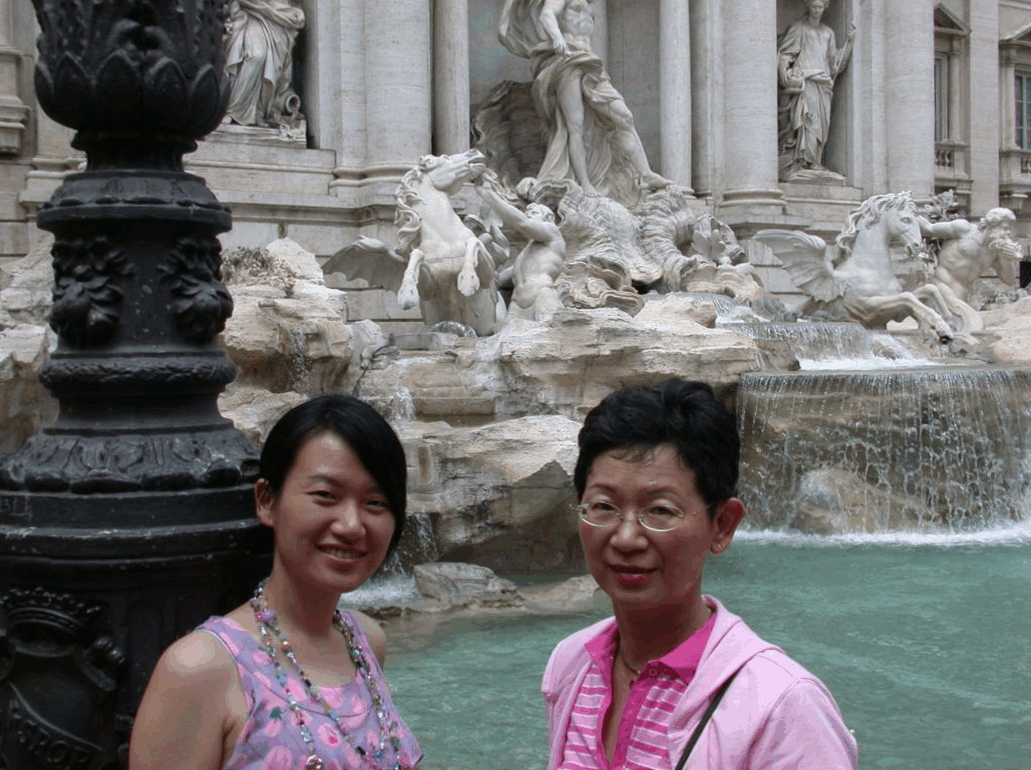} &
         \includegraphics[width=0.24\linewidth]{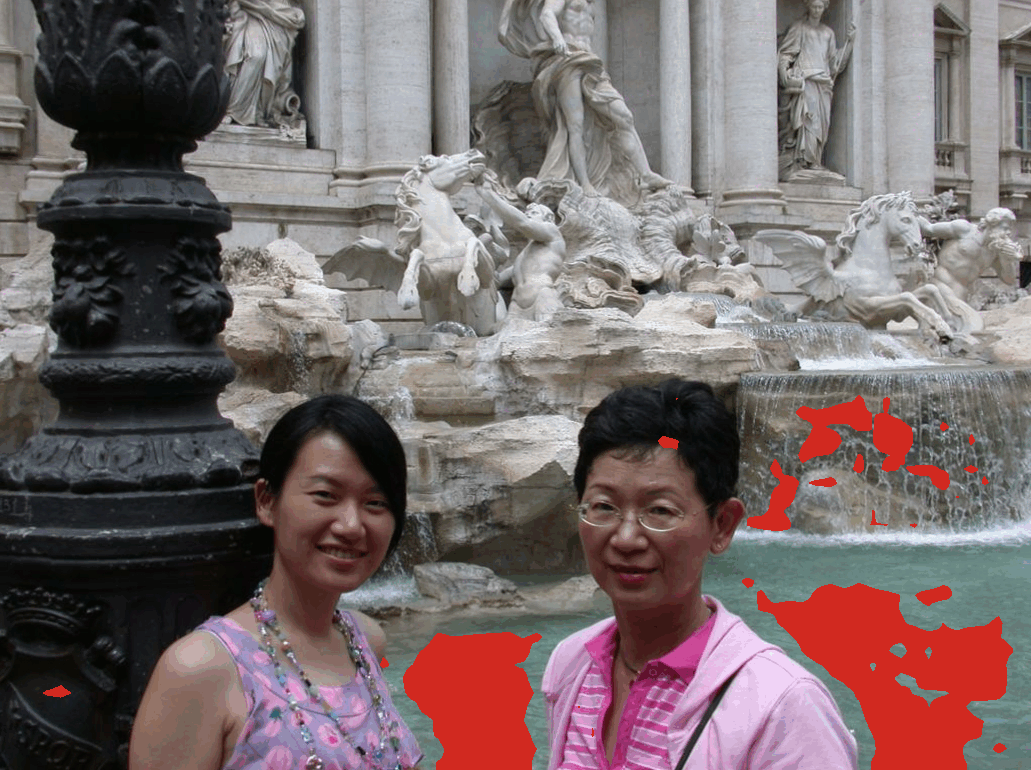} &
         \includegraphics[width=0.24\linewidth]{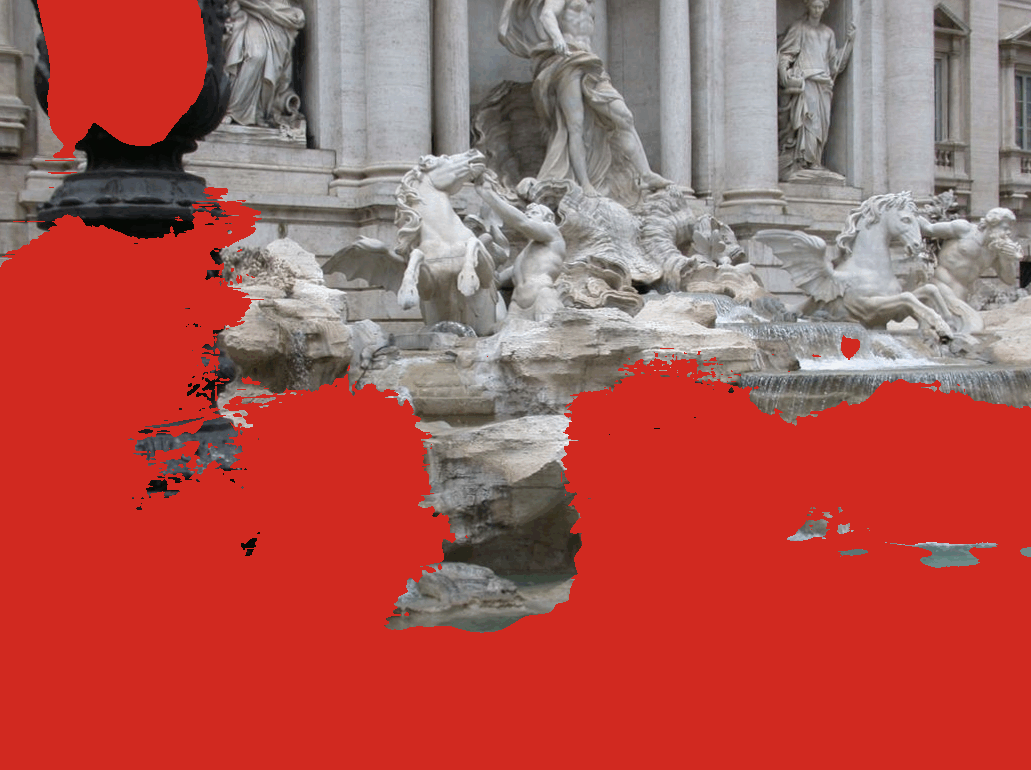} &
         \includegraphics[width=0.24\linewidth]{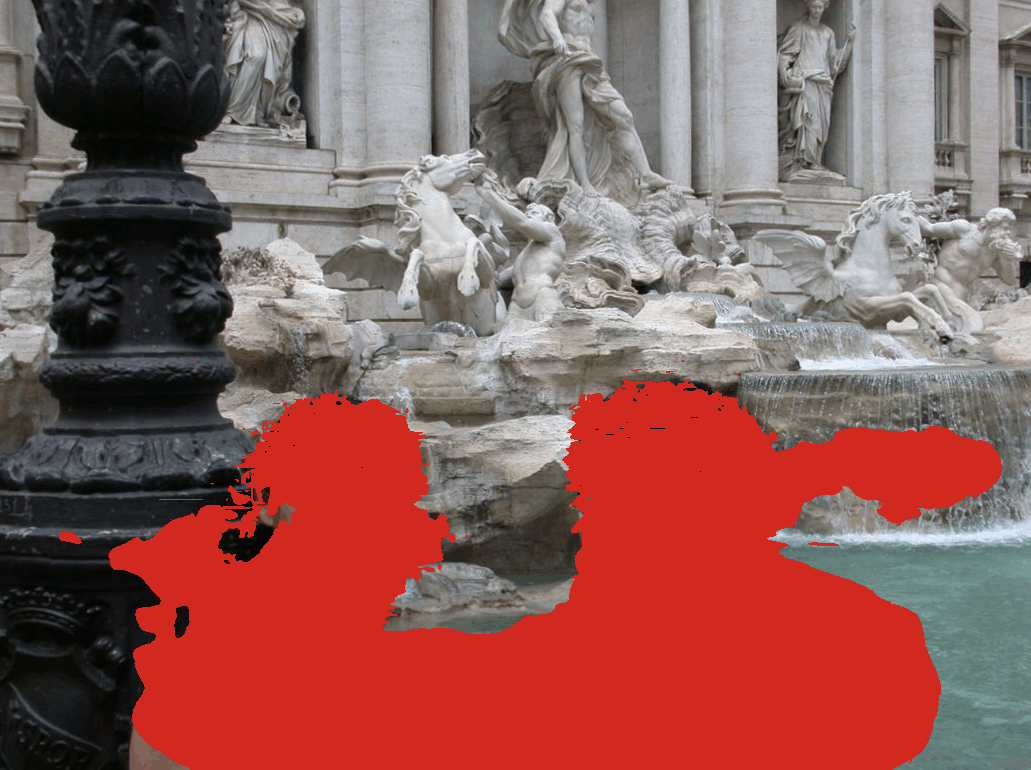} \\
    \end{tabular}
    \vspace{-0.1in}
    \caption{Additional \textbf{Uncertainty Handling Visualization} for three scenes from the Photo Tourism dataset \cite{snavely2006photo}.}
    \label{fig:uncertainty1}
    \vspace{-0.5em}
\end{figure}
%-------------------------------------------------------------------------
\section{Comparisons on NeRF On-the-go}
\label{sec:full results}
While NexusSplats achieves state-of-the-art performance in large-scale outdoor scenes with coupled lighting and occlusion challenges, its quantitative results on the NeRF On-the-Go dataset --- a benchmark focusing on small-scale indoor scenes with graded occlusion levels --- lag behind specialized baselines like WildGauss \cite{kulhanek2024wildgaussians}. This discrepancy stems from two factors inherent to our design priorities and evaluation metrics. 

First, traditional metrics like PSNR, SSIM \cite{wang2004ssim}, and LPIPS \cite{zhang2018unreasonable} measure global photometric fidelity rather than occlusion-specific accuracy. As shown qualitatively in \Cref{fig:quality2}, our method successfully removes transient occlusions (e.g. moving objects in "Patio-High") but incurs slight blurring in boundary regions. NexusSplats prioritizes occlusion robustness over high-frequency detail preservation in small-scale indoor settings. 

Second, the hierarchical representation \cite{lu2024scaffold}, tailored for efficient large-scale scene reconstruction via voxel-aligned kernels, struggles to resolve fine textures in confined indoor spaces. 
This architectural bias toward outdoor scenarios leads to suboptimal parameter tuning for small-scale geometry. Nevertheless, our approach maintains competitive occlusion localization (e.g. 23.41 PSNR in "Corner" vs. WildGauss’s 23.46) while requiring fewer parameters, underscoring its suitability for real-world applications where efficiency and occlusion handling outweigh synthetic benchmark optimization. our future work will explore adaptive kernel scaling to bridge this domain gap without compromising outdoor performance.

%-------------------------------------------------------------------------
\section{Further Ablations and Visualizations}
\label{sec:visualization}
\paragraph{Qualitative Ablations on Photo Tourism.}
We visualize the impact of key components by progressively removing the hierarchical light decoupling module (w/o light.), uncertainty propagation (w/o uncert.), and boundary-aware refinement (w/o refine.).
As shown in \Cref{fig:quality ablation1}, the full NexusSplats framework (rightmost column) restores photometric stability and geometric precision, closely matching ground truth. These visual comparisons underscore the necessity of all three modules for robust \textit{in-the-wild} reconstruction.

\paragraph{Qualitative Ablations on NeRF On-the-go.}
Removing the uncertainty propagation module (w/o uncert.) severely impacts occlusion handling across diverse indoor scenes. In the Corner and Spot sequences, transient objects leave residual artifacts in rendered views, as the model fails to distinguish static structures from occlusions. The Spot scene exhibits blurred boundaries around furniture legs due to unchecked uncertainty accumulation. Similarly, in outdoor-like setups (Mountain and Patio), missing uncertainty propagation learns occlusion geometries and erodes terrain details. The full NexusSplats framework restores precision while preserving intricate textures in partially observed areas. These results validate that our uncertainty propagation mechanism is indispensable for robustly aligning 3D geometry with 2D observations, particularly in scenes with dense transient elements or complex lighting interplay.

\paragraph{Light Decoupling Visualization.}
\Cref{fig:appearance1} and \Cref{fig:appearance2} evaluate the light decoupling performance of NexusSplats on the remaining two scenes in the Photo Tourism dataset \cite{snavely2006photo}, Brandenburg Gate and Sacre Coeur, respectively. The results showcase our method's ability to adapt reconstructed scene colors to match diverse target lighting conditions derived from the ground images. For each scene, six lighting conditions are visualized from three different viewpoints. The consistent adaptation across lighting conditions highlights the effectiveness of the light decoupling module in generalizing across scenes and conditions.

\paragraph{Uncertainty Splatting Visualization.}
\Cref{fig:uncertainty1} presents additional visualization results of uncertainty predictions, highlighting our method's ability to accurately capture occlusions.

%-------------------------------------------------------------------------
\section{Discussions and Limitations}
While NexusSplats significantly improves occlusion handling through 3D uncertainty propagation and boundary-aware refinement, its current formulation exhibits limitations in scenarios requiring fine-grained scene completion. Excluding occluded pixels from the optimization process inherently halts color learning in masked regions, occasionally introducing visible gaps or artifacts in rendered views --- a trade-off inherent to occlusion-filtering frameworks. 

This issue is exacerbated in densely occluded urban scenes, where transient objects mask large contiguous areas, leaving under-optimized "holes" that degrade perceptual quality. A promising remedy lies in integrating image inpainting techniques, such as masked autoencoders \cite{he2022masked} or diffusion models, to hallucinate plausible textures in excluded regions while preserving geometric consistency. 

Additionally, while our hierarchical kernel design excels in large-scale outdoor scenes, it struggles to resolve intricate details in confined indoor environments (e.g. scenes in the NeRF On-the-Go dataset), where high-density representations demand finer-grained spatial partitioning. This limitation stems from the kernel pruning process, which prioritizes computational efficiency over adaptability to varying scene scales.

Future work could explore more robust and scalable kernel optimization to bridge this gap. Despite these challenges, NexusSplats achieves a critical balance: it pioneers efficient, structure-aware occlusion handling without compromising the core advantages of 3DGS --- real-time rendering and photorealistic quality --- in its target domain of \textit{in-the-wild} scene reconstruction.